\pdfoutput=1

\documentclass[lettersize,journal]{IEEEtran}
\usepackage{amsmath,amsfonts}
\usepackage{algorithmic}
\usepackage{array}
\usepackage[caption=false,font=normalsize,labelfont=sf,textfont=sf]{subfig}
\usepackage{textcomp}
\usepackage{stfloats}
\usepackage{url}
\usepackage{verbatim}
\usepackage{graphicx}
\usepackage{balance}

\usepackage{bm}
\usepackage{hyperref}
\usepackage{color}
\usepackage{siunitx}
\usepackage{makecell}

\usepackage{amssymb}
\usepackage[mathscr]{euscript}
\usepackage{comment}
\usepackage{algorithm}
\usepackage{multirow}
\usepackage{wrapfig}
\usepackage{booktabs}
\usepackage{threeparttable}
\usepackage[english]{babel}
\usepackage{microtype}
\usepackage{xcolor}
\usepackage{ulem}

\newcommand{\etal}{\textit{et al.}}

\newcommand{\hl}[1]{\textcolor{red}{#1}}

\newcommand{\para}[1]{\noindent\textbf{#1}}

\def\BibTeX{{\rm B\kern-.05em{\sc i\kern-.025em b}\kern-.08em
    T\kern-.1667em\lower.7ex\hbox{E}\kern-.125emX}}

\begin{document}
	
\title{HeadRecon: High-Fidelity 3D Head Reconstruction from Monocular Video}
% NNR-Head: Neural Non-Rigid 3D Head Reconstruction from Monocular Videos

\author{Xueying~Wang and Juyong~Zhang$^{\dagger}$
\thanks{Xueying Wang is with the Department of Basic Courses, Naval University of Engineering.}
\thanks{Juyong Zhang is with the School of Mathematical Sciences, University of Science and Technology of China.}
\thanks{$^\dagger$Corresponding author. Email: {\texttt{juyong@ustc.edu.cn}}.}}

% \markboth{Submitted to IEEE Transactions on Image Processing}%
% {Wang \MakeLowercase{\textit{et al.}}: NNR-Head: Neural Non-Rigid 3D Head Reconstruction from Monocular Videos}

\maketitle

\begin{abstract}
Recently, the reconstruction of high-fidelity 3D head models from static portrait image has made great progress. However, most methods require multi-view or multi-illumination information, which therefore put forward high requirements for data acquisition. In this paper, we study the reconstruction of high-fidelity 3D head models from arbitrary monocular videos. Non-rigid structure from motion (NRSFM) methods have been widely used to solve such problems according to the two-dimensional correspondence between different frames. However, the inaccurate correspondence caused by high-complex hair structures and various facial expression changes would heavily influence the reconstruction accuracy. To tackle these problems, we propose a prior-guided dynamic implicit neural network. Specifically, we design a two-part dynamic deformation field to transform the current frame space to the canonical one. We further model the head geometry in the canonical space with a learnable signed distance field (SDF) and optimize it using the volumetric rendering with the guidance of two-main head priors to improve the reconstruction accuracy and robustness. Extensive ablation studies and comparisons with state-of-the-art methods demonstrate the effectiveness and robustness of our proposed method.
\end{abstract}

\begin{IEEEkeywords}
Non-rigid 3D head reconstruction, dynamic deformation field, implicit volumetric rendering, prior guidance
\end{IEEEkeywords}

\IEEEpeerreviewmaketitle

\begin{figure*}[htbp]
    \centering
    \includegraphics[width=0.99\textwidth]{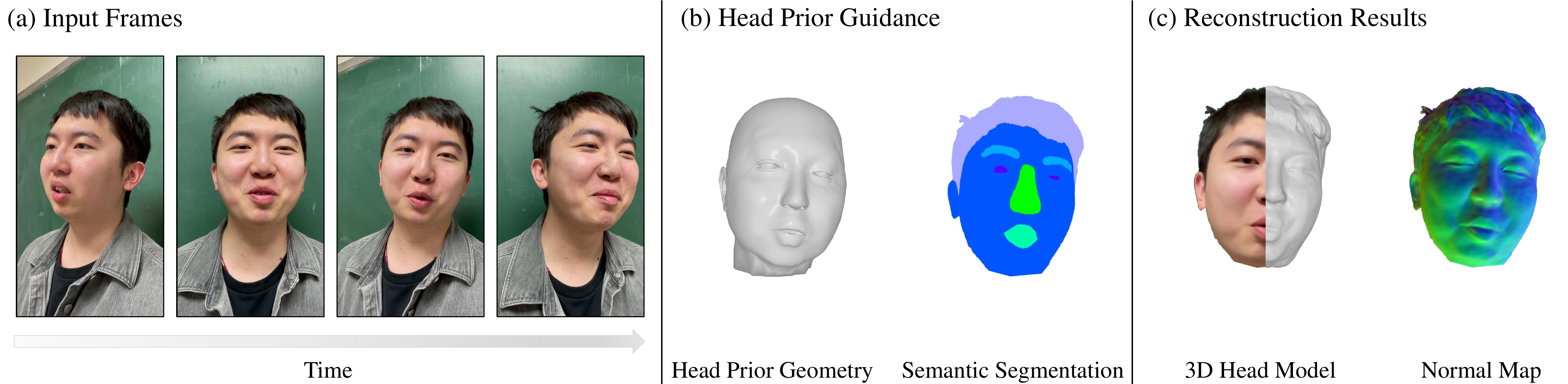}
    \caption{Given a monocular dynamic video sequence (a), our method obtains two main head priors for each frame, including the head prior geometry and semantic segmentation information (b). With the guidance of this prior knowledge, we can reconstruct high-fidelity head models frame-by-frame (c) via a dynamic implicit volumetric rendering framework.}
    \label{fig:teaser}
\end{figure*}

\section{Introduction}

\IEEEPARstart{H}igh-fidelity 3D head reconstruction has been a research hot spot in computer vision and computer graphics due to its wide applications in the game character design \cite{shi2019face,shi2020neutral}, film production \cite{song2020accurate,lightstage,zhou20183d,hong2022headnerf}, and human-computer interaction  \cite{anbarjafari20173d,lifkooee2018image,DBLP:journals/tmm/LouWNHMWY20}. Such a high-fidelity 3D head model can be obtained directly by scanning a static person with a fixed pose and expression, but this approach requires expensive equipment like laser scanners and may result in noisy and incomplete models. Meanwhile, image-based reconstruction methods have achieved faithful results. These methods could recover high-fidelity 3D geometries from static portrait images under multi-view \cite{9537697,zhang2022video,li2021topologically} or multi-illumination \cite{wang2020lightweight,cao2018sparse,chen20203d,debevec2000acquiring} conditions. However, they heavily rely on a synchronized data acquisition process with the sophisticated camera and lighting settings. Besides, the collection process may be perturbed by the unconscious pose and expression changes of the collected person. All these shortcomings seriously limit their applicability in daily life. Meanwhile, with the popularity of portable devices and the Internet, we can easily obtain a dynamic monocular video of a person from different sources, such as TV programs and social media. Thus, to reduce data acquisition requirements, we focus on reconstructions from dynamic videos and aim to reconstruct frame-by-frame head models including the hair region from arbitrary monocular dynamic video sequences.

% application prospects
% These methods usually take static portrait images as input, and use multi-view \cite{9537697,zhang2022video,li2021topologically} or multi-illumination \cite{wang2020lightweight,cao2018sparse,chen20203d,debevec2000acquiring} cues to recover high-fidelity 3D geometries.
% To reduce data acquisition requirements, we focus on non-rigid structure from motion methods and aim to reconstruct frame-by-frame integrated head models including the hair region from monocular dynamic video sequences.
% with pose and expression changes.

%\hlb{To model non-rigid deformations in the dynamic videos, we pay attention to Non-Rigid Structure from Motion (NRSFM) methods. NRSFM} 

Non-Rigid Structure from Motion (NRSFM) \cite{jensen2021benchmark} has been widely adopted in the reconstruction of non-rigid surfaces from monocular video sequences according to the motion and deformation cues. Its reconstruction accuracy heavily depends on the accuracy of correspondences between different frames. Unfortunately, most related works only use sparse feature points as correspondent trails \cite{kong2016prior,dai2014simple,zhu2014complex}. Even though dense NRSFM methods \cite{Sidhu2020,ansari2017scalable,golyanik2019consolidating,golyanik2017dense} have made great progress, they still suffer from complex geometry modeling and data pre-processing to construct the dense correspondence. Furthermore, in the specific field of non-rigid 3D head reconstructions, it is difficult to extract accurate explicit correspondence owing to the highly complex structures of hair regions and various facial expression changes. Thus, most related NRSFM methods are only able to model the coarse facial geometry \cite{bai2020deep,bai2021riggable}.

% \hlb{Traditional optimization-based methods generally profit from the low-rank geometry assumption.}
% The specific field of 3D head reconstructions is benefited from the basic geometry structure of the human head. Most existing works combine the parametric head models \cite{blanz1999morphable,FLAME:SiggraphAsia2017} with NRSFM and use deep learning-based methods \cite{bai2020deep,bai2021riggable} to avoid the complex correspondence extraction. Nevertheless, they are data-hungry and are only able to model the coarse facial geometry \hlb{without the hair region.}

To avoid the inaccurate extractions of the 2D correspondence, some methods adopt the self-supervised learning to solve the dynamic reconstruction problem via the photometric consistency. Some approaches employ the differentiable rendering based on template models to reconstruct dynamic models. Recently, a differentiable rendering-based 3D head reconstruction approach for dynamic monocular videos has been proposed in \cite{grassal2022neural}, which first recovers the basic geometry frame-by-frame with a parametric head model and then refines the geometry shape with the vertex-wise optimizations. However, it is still limited by the expression ability of the parametric model and its performance may fluctuate on different input dynamic video sequences. On the other hand, methods combining the implicit representation and volumetric rendering \cite{mildenhall2020nerf} have attracted a lot of attentions. These methods benefit from both the flexibility of the implicit representation and the authenticity of the volumetric rendering to be capable of modeling an arbitrary single scene. Moreover, dynamic deformation fields are integrated with such methods for the non-rigid deformation modeling\cite{tretschk2021non,pumarola2021d}. However, these methods cannot be directly used for non-rigid deformation head reconstructions, mainly for the following reasons. Firstly, it is difficult to design a proper dynamic deformation approach to model the non-rigid deformations. Secondly, most methods generally focus on the novel view synthesis and adopt the density-based representation, making the reconstructed geometric surfaces noisy and rough. Finally, the self-supervised learning strategy based on the photometric consistency may lead to depth ambiguity.

% To avoid the requirement for accurate 2D correspondence extractions,
% Some approaches try to reconstruct the dynamic head models via the parametric model\cite{grassal2022neural}. They first recover the basic geometry frame-by-frame and then refine the detailed information with the vertex-wise optimizations. However, they are still limited by the expressive ability of the template model and their performance may fluctuate on different input dynamic video sequences. Recently, methods combining the implicit representation and volumetric rendering \cite{mildenhall2020nerf} have attracted a lot of attentions. These methods benefit from both the flexibility of the implicit representation and the authenticity of the volumetric rendering to be capable of modeling an arbitrary single scene. Furthermore, dynamic deformation fields are increasingly involved in such methods for the non-rigid deformation modeling \cite{tretschk2021non,pumarola2021d}. Nevertheless, these methods are still not straightforward for non-rigid deformation head reconstructions, mainly for the following reasons. Firstly, it is difficult to design a proper dynamic deformation approach to model the non-rigid deformations. Secondly, most methods generally focus on the novel view synthesis and adopt the density-based representation, making the reconstructed geometric surfaces noisy and rough. Finally, the self-supervised learning strategy based on the photometric consistency may lead to depth ambiguity.
% focus on the novel view synthesis with high-fidelity rendering results and

%with pose and expression changes
In this paper, we combine both the dynamic implicit volumetric rendering and head priors to reconstruct frame-by-frame high-fidelity head models from monocular dynamic video sequences. To properly model non-rigid deformations, we split the dynamic deformation field into two parts by first rigidly registering the current frame space into the canonical one, \iffalse\hl{(no pose and expression?)}\fi and then computing the point-wise offsets. Meanwhile, we adopt an ambient module to describe the topology features \cite{park2021hypernerf}. Moreover, to reduce the depth ambiguity and geometric outliers, we utilize the head prior geometry and semantic segmentation information as constraints to improve the reconstruction robustness and accuracy. Specifically, we first adopt an optimization-based approach to recover frame-by-frame coarse 3D FLAME models \cite{FLAME:SiggraphAsia2017} and camera parameters from input video sequences. This information provides reasonable 3D structures and an initial depth range for the implicit space. \iffalse\hl{(coarse deformations?)}\fi Then, we employ the 2D semantic segmentation information obtained with face parsing model \cite{lee2020maskgan} to maintain the reconstruction accuracy. Using this auxiliary information, we estimate the head geometry in the canonical space as a zero-level set of the signed distance field represented using an implicit volumetric rendering network \cite{wang2021neus}. At last, frame-by-frame high-fidelity 3D head models are recovered via Marching Cube \cite{lorensen1987marching}. 

Fig.~\ref{fig:teaser} shows the overview of our method. Furthermore, we can also achieve facial reenactment via our proposed dynamic reconstruction framework due to the disentangled expression representation ability provided by FLAME models \cite{FLAME:SiggraphAsia2017}. Comprehensive experiments demonstrate that our method can reconstruct more accurate head models from monocular dynamic video sequences, compared with state-of-the-art head reconstruction methods from monocular videos. In summary, the contributions of this work include the following aspects:

\begin{itemize}
    \item We propose a prior-guided dynamic implicit neural network structure for the reconstructions of frame-by-frame high-fidelity 3D head models from monocular dynamic video sequences.
    % with pose and expression changes.
    
    \item We combine the overall rigid registration and point-wise non-rigid displacement to model the dynamic non-rigid deformations. Two head priors including the head prior geometry and semantic segmentation information are introduced to further improve the reconstruction accuracy and robustness.
    
    \item With our proposed algorithm framework, facial reenactment can be achieved, implying the potentiality of our method to digital human head-related applications.
    
\end{itemize}

% \hl{non-rigid head reconstruction methods}

% We introduce two main head priors to improve the reconstruction accuracy and robustness. The head prior geometry provides the initial geometric information and depth range for the implicit space. And the head semantic segmentation information helps to maintain the geometric accuracy.

% \item Extensive experiments demonstrate that our method performs well on various types of monocular dynamic video sequences.
\section{Related Work}

In this section, we review some relevant literature from four aspects: traditional and deep learning-based non-rigid structure from motion, head reconstruction from video sequences, and dynamic radiance fields.

\textbf{Non-Rigid Structure from Motion.} The non-rigid structure from motion method recovers non-rigid surfaces from monocular image sequences according to the motion and deformation cues. Therefore, it is important to establish accurate two-dimensional correspondence from multi-frames for reconstruction accuracy. In traditional optimization-based methods, it is a common way to construct low-rank geometric assumptions as priors. Bregler \etal~\cite{bregler2000recovering} employed a factorization approach to recover non-rigid surfaces by representing the per-frame geometry as a linear combination of a set of basis. Dai \etal~\cite{dai2014simple} directly imposed the low-rank geometric assumptions on a time-varying shaped matrix via the trace norm minimization. Zhu \etal~\cite{zhu2014complex} modeled the relatively complex human motion as a union of shapes to reconstruct body skeletons. However, sparse features used in these works would affect the reconstruction accuracy, so dense features were introduced. Agudo \etal~\cite{agudo2017global} optimized the global geometry while tuning its contributions to local regions divided by their physical behaviors. Garg \etal~\cite{garg2013dense} formulated the reconstruction as a dense variational problem and solved it according to dense long-term 2D trajectories. While Kumar \etal~\cite{kumar2018scalable} used the Grassmann manifold to model the non-rigid deformation spatially and temporally based on a union of linear subspaces. Even though the dense correspondence has a positive effect on the reconstruction accuracy, a more complex optimization process is also introduced. Thus, we use a deep learning-based method to avoid such complex optimizations.

\textbf{Deep Learning-Based Non-Rigid Structure from Motion.} With the development of convolutional neural networks, deep learning-based non-rigid structure from motion methods have made great progress. Kong \etal~\cite{kong2019deep} proposed a novel deep neural network to jointly recover camera poses and 3D points by learning a series of dictionaries end-to-end. While Novotny \etal~\cite{novotny2019c3dpo} used a factorization network to estimate the viewpoint information and shape parameters based on the 2D key point information through a canonicalization loss. However, these works are based on supervised learning, and trained on the amount of data with 3D ground truth or accurate 2D annotations, resulting in limited applications. To enable unsupervised learning, Sidhu \etal~\cite{Sidhu2020} designed an auto-encoder to formulate the deformation model and subspace constraints to regularize the recovered latent space function. Takmaz \etal~\cite{takmaz2021unsupervised} presented an unsupervised monocular framework to learn dense depth maps by preserving pairwise distances between reconstructed 3D points. Nevertheless, the use of the explicit dense correspondence, especially the optical flow, would lead to either complex network structures and loss function designs or data preprocessing procedures. So in this paper, we learn the non-rigid motions and deformations directly through implicit dynamic deformation fields to circumvent the explicit correspondence constructions.

\textbf{Face Reconstruction from RGB videos.} Various methods have been proposed to solve the problem of 3D face reconstructions from RGB videos in recent years. Most existing works introduced 3D parametric face models to solve this problem~\cite{lou2021real,wu2021f3a}. Garrido \etal~\cite{garrido2013reconstructing} used a blend shape model to capture the face geometry by tracking accurate sparse 2D features between selected key frames, and further refined it based on the temporally coherent optical flow and photometric stereo. They further proposed an approach for the automatic creation of a personalized high-quality 3D face rig of an actor to reconstruct facial details, like folds and wrinkles, in a multi-scale capture manner \cite{garrido2016reconstruction}. While most deep learning-based works attempted to regress the coefficients of the 3D Morphable Model (3DMM) \cite{blanz1999morphable,paysan20093d,booth20163d} from the input RGB sequence. To maintain the identity consistency between different frames, Deng \etal~\cite{deng2019accurate} designed a confidence measurement network. The combination of multi-level feature maps and the multi-view appearance consistency was adopted in \cite{bai2020deep} to gradually refine facial geometries. Guo \etal~\cite{guo2018cnn} used a coarse-to-fine strategy to first reconstruct the basic identity and expression information and then recover the facial details. They further proposed a more powerful convolutional network to directly decode the facial geometry from encoded features via the photometric consistency \cite{guo20213d}.

However, almost all these works only focus on the face region, and still cannot meet the requirements of high-precision reconstructions due to the limited expressive ability of the parametric face models. Recently, there have been some works devoted to the reconstruction of dynamic head models. Grassal \etal~\cite{grassal2022neural} first regressed a head parametric model from an input RGB frame, then computed per-vertex offsets to recover the hair region and refine the face region. While Zheng \etal~\cite{zheng2021avatar} used an implicit network to learn the head geometry and represented the deformations via learned blendshapes and skinning fields, a similar idea was also adopted in \cite{jiang2022selfrecon}. Different from their works, we can directly calculate the non-rigid deformation for a given 3D point, instead of establishing the parametric model-based displacements or the correspondence searching procedures.

\textbf{Dynamic Radiance Fields.} Once proposed, the neural radiance field (NeRF) \cite{mildenhall2020nerf} has received extensive attention, and it has a desirable performance in static scenes. Many follow-up works have made a great effort to model dynamic scenes, so-called the dynamic radiance field. Gafni \etal~\cite{gafni2021dynamic} introduced a low-dimensional morphable model to provide explicit controllable pose and expression codes, and modeled the dynamic performance together with learnable latent codes. Meanwhile, there were some works describing the non-rigid deformations with additional dynamic deformation fields. They generally encoded the dynamic scene into per-frame current spaces and the shared canonical space, and established their maps through the dynamic deformation fields. Given a 3D point at a specific time instant, Pumarola \etal~\cite{pumarola2021d} computed its 3D displacement from the current space to the canonical one. Tretschk \etal~\cite{tretschk2021non} implemented the displacements as the ray bending which meant camera rays in the current space can be non-rigidly deformed, and proposed a rigidity network to better constrain the rigid regions. Both rotations and translations of a given 3D point were estimated in \cite{mildenhall2020nerf}, and an elastic regularization was introduced to improve the robustness. To better illustrate the topology changes during deformations, a higher dimension was further introduced in \cite{park2021hypernerf} as the ``ambient'' space. Since these works commit to doing the novel view synthesis, they obtain the high-fidelity rendering results at the cost of low-fidelity geometries. In this paper, we adopt the signed distance field for the geometric representation \cite{wang2021neus}, and introduce head priors to further improve the reconstruction accuracy and robustness. 

\begin{figure*}[htbp]
    \centering
    \includegraphics[width=1.0\textwidth]{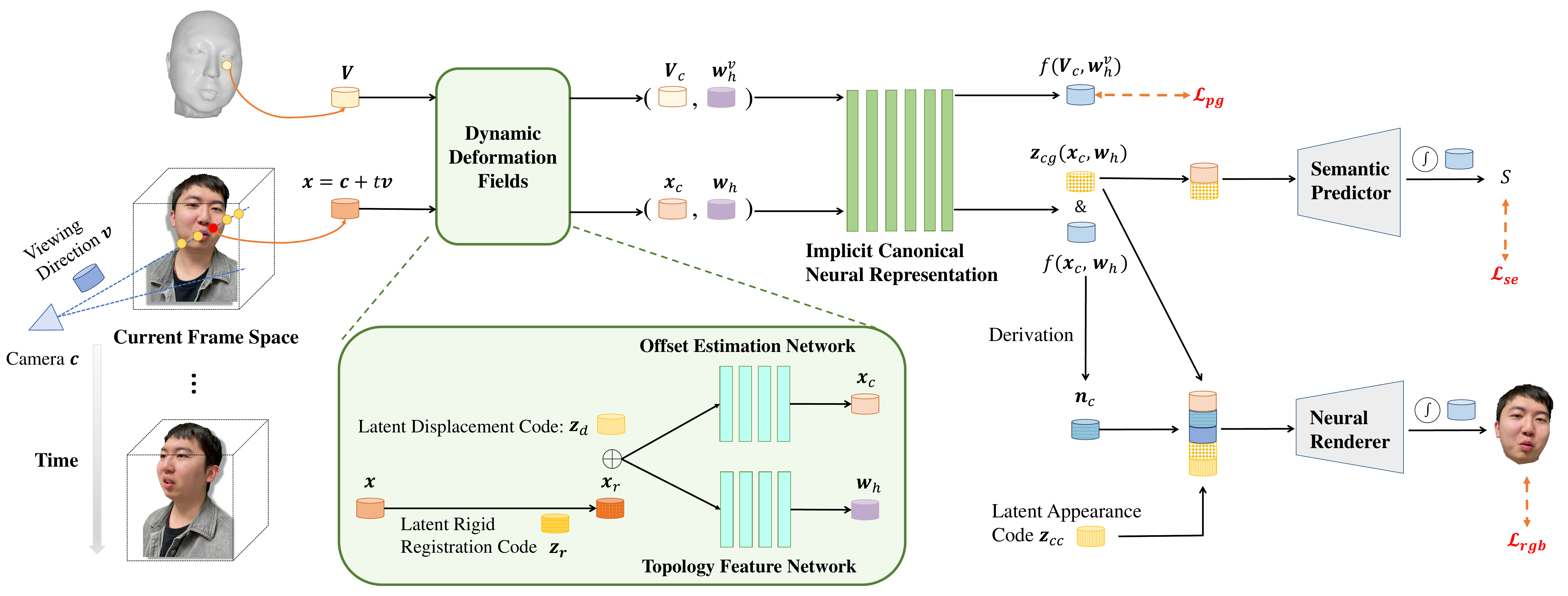}
    \caption{Our proposed algorithm pipeline. We design a two-part dynamic deformation field to model per-frame non-rigid deformations, and reconstruct the 3D head models frame by frame via the implicit volumetric rendering method mainly according to the photometric consistency $\mathcal{L}_{rgb}$. Meanwhile, we construct two main head priors for each input frame, including the head prior geometry and semantic segmentation information. Then we employ two prior-guided loss terms $\mathcal{L}_{pg}$, $\mathcal{L}_{se}$ in our framework to improve the reconstruction accuracy and robustness.}
    \label{fig:pipeline}
\end{figure*}

% Our proposed algorithm pipeline. We design a two-part dynamic deformation field to model per-frame non-rigid deformations, and construct two main head priors for each input frame, including the head prior geometry and semantic segmentation information. Firstly, self-supervision signals derive from the difference between the input frame and the rendered frame. Meanwhile, we further employ two prior-guided loss terms in our framework to improve the reconstruction accuracy and robustness. Finally, we reconstruct the 3D head models frame by frame via the implicit volumetric rendering method.
\section{Approach}

%with pose and expression changes. 
In this section, we represent our method design that reconstructs head models including the hair regions frame-by-frame from monocular dynamic video sequences. When restricted to modeling dynamic non-rigid deformations, previous works suffer from the coarse facial geometry and depth ambiguity due to the low-rank geometry assumption or the density-based representation, and the self-supervised learning strategy based on the photometric consistency. To tackle these problems, we construct a two-part deformation field (Sec.~\ref{sec:ddf}) and introduce two main head priors, including the head prior geometry (Sec.~\ref{sec:proxy}) and semantic segmentation information (Sec.~\ref{sec:semantic}). Furthermore, benefiting from the flexibility of the implicit representation and the authenticity of the volumetric rendering (Sec.~\ref{sec:ivrn}), we propose a prior-guided dynamic implicit volumetric rendering network with carefully designed loss terms (Sec.~\ref{sec:loss}) to improve the reconstruction accuracy and robustness. We show the pipeline of our proposed method in Fig.~\ref{fig:pipeline}, and describe each component in detail in the following subsections.

\subsection{Preliminaries.}

\textbf{Parametric Head Model.} The FLAME head model \cite{FLAME:SiggraphAsia2017} describes the head shape, pose and facial expression in a low-dimensional parametric subspace. In this paper, we only use this parametric model to represent the coarse shape and expressions of a given subject. Thus the simplified FLAME head model can be described as follows:
\begin{equation}
    \begin{aligned}
    \label{eq:flame_geo}
    \mathbf{G} = W(T(\bm{\alpha}_{id}, \bm{\alpha}_{exp}), J(\bm{\alpha}_{id}), \mathbf{W}) \in \mathbb{R}^{3n_v}, 
    \end{aligned}
\end{equation}
where $n_v$ is the vertex number of the head model. $\bm{\alpha}_{id} \in \mathbb{R}^{300}$,  and $\bm{\alpha}_{exp} \in \mathbb{R}^{100}$ are shape and expression parameters respectively. $W(\cdot)$ is the standard skinning function. $J(\cdot)$ is the joint regressor, and $\mathbf{W} \in \mathbb{R}^{k \times n_v}$ is the skinning weight matrix for smooth blending based on pre-defined $k$ joints. $T(\bm{\alpha}_{id}, \bm{\alpha}_{exp})$ denotes the template with added shape and expression offsets:
\begin{equation}
    \label{eq:flame_tgeo}
    T(\bm{\alpha}_{id}, \bm{\alpha}_{exp}) = \overline{\mathbf{T}} + B_S(\bm{\alpha}_{id}; \mathbf{\mathcal{S}}) + B_E(\bm{\alpha}_{exp}; \mathbf{\mathcal{E}}), 
\end{equation}
where $\overline{\mathbf{T}} \in \mathbb{R}^{3n_v}$ is the template model. $B_S(\cdot)$ and $B_E(\cdot)$ calculate variations in the shape and expression using their corresponding linear blendshapes $\mathbf{\mathcal{S}}$ and $\mathbf{\mathcal{E}}$ respectively, conditioned on given $\bm{\alpha}_{id}$ and $\bm{\alpha}_{exp}$. This parametric model also constructs the head texture map $\mathbf{A} \in \mathbb{R}^{512 \times 512 \times 3}$ with principal component analysis (PCA) as \cite{flametexture}:
\begin{equation}
    \label{eq:flame_tex}
    \mathbf{A} = \overline{\mathbf{A}} + \mathbf{B}_{tex}\bm{\alpha}_{tex}, 
\end{equation}
where $\overline{\mathbf{A}} \in \mathbb{R}^{512 \times 512 \times 3}$ is the mean texture map. $\mathbf{B}_{tex} \in \mathbb{R}^{512 \times 512 \times 3 \times 50}$ is the principal axes extracted from the FFHQ dataset \cite{karras2019style} using a model fitting method, and $\bm{\alpha}_{tex} \in \mathbb{R}^{50}$ is the corresponding coefficient parameters. We use the textured FLAME to build a head proxy model for each frame, and treat it as the head prior geometry for further network training.

\textbf{Camera Model.} We use the standard perspective projection to project a 3D head model onto a 2D image plane, which can be expressed as follows:
\begin{equation}
    \label{eq:camera}
    \mathbf{q} = \bm{\Pi}(\mathbf{R}\mathbf{V} + \mathbf{t}),
\end{equation}
where $\mathbf{q} \in \mathbb{R}^2$ is the location of vertex $\mathbf{V} \in \mathbb{R}^3$ in the image plane. $\mathbf{t} \in \mathbb{R}^3$ is the translation vector, and $\mathbf{R} \in \mathbb{R}^{3 \times 3}$ is the rotation matrix constructed from Euler angles $pitch, yaw$ and $roll$. $\bm{\Pi}: \mathbb{R}^3 \rightarrow \mathbb{R}^2$ is the perspective projection. For convenience, we denote a group of camera parameters as $\bm{\tau} = \left\{pitch, yaw, roll, \mathbf{t}\right\}$ and compute the camera center as $\mathbf{c} = \mathbf{c}(\bm{\tau}) = -\mathbf{R}^{T}\mathbf{t}$.

\subsection{Dynamic Deformation Fields}
\label{sec:ddf}

We model the non-rigid deformations with a two-part dynamic deformation field. We first rigidly register the current frame space into the canonical one using a SE(3) mapping function as follows:
\begin{equation}
    \label{eq:dr_field}
    \begin{aligned}
    D_r: \mathbb{R}^{6} &\rightarrow \mathbb{R}^{3 \times 3}, \mathbb{R}^{3} \\
    \mathbf{z}_r &\mapsto \mathbf{R}_r, \mathbf{t}_r
    \end{aligned}
\end{equation}
where $\mathbf{z}_r \in \mathbb{R}^{6}$ is a per-frame trainable latent code that encodes the state of the head model in each frame. $\mathbf{R}_r \in \mathbb{R}^{3 \times 3}$ and $\mathbf{t}_r \in \mathbb{R}^3$ are the rotation matrix and translation vector of the rigid registration respectively, obtained via the Lie Algebra. Then we implicitly estimate the point-wise offset utilizing a multi-layer perceptron (MLP) $d$ with trainable parameters $\bm{\beta} \in \mathbb{R}^{n_d}$:
\begin{equation}
    \label{eq:do_field}
    \Delta\mathbf{\mathcal{D}}_{\bm{\beta}}(\mathbf{x}, \mathbf{z}_d) = d\left( \mathbf{x}, \mathbf{z}_d \ | \ \bm{\beta} \right),
\end{equation}
where $\mathbf{x} \in \mathbb{R}^{3}$ is a point in the 3D space, and $\mathbf{z}_d \in \mathbb{R}^{100}$ encodes the per-frame displacements which is set as the FLAME expression parameter $\bm{\alpha}_{exp}$. Specifically, for a given pixel in the current frame $F$, indexed by $p$, let $\mathbf{c}_p = \mathbf{c}_p(\bm{\tau})$ represent the center of the corresponding camera, with $\mathbf{v}_p = \mathbf{v}_p(\bm{\tau})$ as the viewing direction (i.e. the vector pointing from $\mathbf{c}_p$ to $p$). Then, $R_p=\left\{ \mathbf{c}_p + t\mathbf{v}_p | t \geq 0 \right\}$ represents the ray through pixel $p$. Let $\mathbf{x}_p = \mathbf{x}_p(t)$ be the sampled point along the ray $R_p$, then its corresponding position $\mathbf{x}_r = \mathbf{x}_r(t)$ after the rigid registration can be expressed as:
\begin{equation}
    \label{eq:x_r}
    \mathbf{x}_r = \mathbf{R}_r\mathbf{x}_p + \mathbf{t}_r,
\end{equation}
further, its corresponding canonical position $\mathbf{x}_c = \mathbf{x}_c(t)$ can be obtained by:
\begin{equation}
    \label{eq:x_c}
    \mathbf{x}_c = \mathbf{x}_r + \Delta\mathbf{\mathcal{D}}_{\bm{\beta}}\left( \mathbf{x}_r, \mathbf{z}_d \right).
\end{equation}
Meanwhile, inspired by \cite{park2021hypernerf}, to describe the topology changes we use another MLP $h$ to predict the ambient coordinate $\mathbf{w}_h \in \mathbb{R}^{2}$ of $\mathbf{x}_r$ conditioned on $\mathbf{z}_d$ with trainable parameters $\bm{\gamma} \in \mathbb{R}^{n_h}$. Different from \cite{park2021hypernerf}, we use the implicit signed distance field to model the head geometry to enable the high-fidelity reconstruction.

\subsection{Implicit Volumetric Rendering Network}
\label{sec:ivrn}

We implicitly model the head geometry in the canonical space with an MLP $f$ that estimates the signed distance of a given point with its ambient coordinate. Under such a formulation, the head geometry can be extracted as the zero-level set of the neural network $f$:
\begin{equation}
    \label{eq:sdf}
    \mathbf{\mathcal{H}}_{\bm{\beta, \gamma, \eta}} = \left\{ \mathbf{x}_c \in \mathbb{R}^3 \ | \ f(\mathbf{x}_c, \mathbf{w}_h | \bm{\eta}) = 0 \right\},
\end{equation}
where $\bm{\eta} \in \mathbb{R}^{n_f}$ is the trainable network parameters. To enable the differentiable rendering of the head region, we use the volumetric rendering to predict the appearance of the camera ray $R_p$, inspired by \cite{wang2021neus}. We first obtain the rendered color of point $\mathbf{x}_c$ via a MLP $g$ with trainable parameters $\bm{\theta} \in \mathbb{R}^{n_g}$, which can be expressed as:
\begin{equation}
    \label{eq:sampled_rgb}
    \mathbf{\mathcal{A}}_{\bm{\beta, \gamma, \eta, \theta}}(\mathbf{x}_c) = g( \mathbf{x}_c, \mathbf{n}_c, \mathbf{v}_c, \mathbf{z}_{cg}, \mathbf{z}_{cc} \ | \ \bm{\theta}),
\end{equation}
where $\mathbf{z}_{cc} \in \mathbb{R}^{64}$ decodes the appearance information in the current frame $F$. $\mathbf{n}_c \in \mathbb{R}^3$ is the normal vector of $\mathbf{x}_c$ derived as the gradient of $f$ at $\mathbf{x}_c$. $\mathbf{v}_c \in \mathbb{R}^3$ is the modified viewing direction in the canonical space, and $\mathbf{z}_{cg} \in \mathbb{R}^{256}$ is the global geometry feature vector, both of them are described in Sec.~\ref{sec:implement} Then, we can obtain the rendered appearance of the camera ray $R_p$ as follows:
\begin{equation}
    \label{eq:rgb}
    \mathbf{\mathcal{C}}_{\bm{\beta, \gamma, \eta, \theta}}(p) = \int_{t_n}^{t_f}T_{sm}(t)\sigma(\mathbf{x}_c(t))\mathbf{\mathcal{A}}_{\bm{\beta, \gamma, \eta, \theta}}(\mathbf{x}_c(t))dt,
\end{equation}
where $t_n$, $t_f$ are the near and far steps. $\sigma(\mathbf{x}_c(t))$ denotes the volume density extracted from the signed distance field via the sigmoid function $\Phi_{sig}(x) = (1 + e^{-sx})^{-1}$:
\begin{equation}
    \label{eq:density}
    \sigma(\mathbf{x}_c(t)) = \max \left( \displaystyle\frac{-\frac{d\Phi_{sig}}{dt}(f(\mathbf{x}_c(t), \mathbf{w}_h))}{\Phi_{sig}(f(\mathbf{x}_c(t), \mathbf{w}_h))}, 0 \right),
\end{equation}
$T_{sm}(t)$ is the accumulated transmittance along $R_p$ from $t_n$ to $t$ which can be expressed as follows:
\begin{equation}
    \label{eq:transmittance}
    T_{sm}(t) = exp(-\int_{t_n}^{t}\sigma(\mathbf{x}_c(u))du).
\end{equation}
Based on this implicit volumetric rendering, we further use the head prior geometry and semantic segmentation information extracted from monocular dynamic video sequences to improve the reconstruction accuracy and robustness.

\subsection{Head Proxy Model}
\label{sec:proxy}

To avoid the unreasonable head structure and depth ambiguity, we introduce the head prior geometry to guide the head reconstruction outside the hair region. We refer to FLAME \cite{FLAME:SiggraphAsia2017}, a widely used 3D parametric head model, for the head prior geometry. Given $N$ input frames extracted from a monocular dynamic video sequence, we employ an optimization-based inverse rendering method to estimate the FLAME parameters $(\bm{\alpha}_{id}, \bm{\alpha}_{exp}, \bm{\alpha}_{tex})$, camera parameters $\bm{\tau}$ and lighting conditions using the photometric consistency. We assume that the human head outside the hair region can be regarded as a Lambertian surface, and use spherical harmonics (SH) basis functions to simulate the global illumination. Under this assumption, the imaging formula for a vertex indexed by $i$ can be expressed as:
\begin{equation}
    \label{eq:image_formula}
    \mathbf{I}(\mathbf{n}_i, \mathbf{b}_i \ | \ \bm{\nu}) = \mathbf{b}_i \cdot (\bm{\nu} \cdot \bm{\phi}(\mathbf{n}_i)),
\end{equation}
where $\mathbf{b}_i \in \mathbb{R}^3$ is the albedo in the RGB color space, $\bm{\phi}(\mathbf{n}_i) \in \mathbb{R}^{B^2}$ is the SH basis function computed with the vertex normal $\mathbf{n}_i$, and $\bm{\nu} \in \mathbb{R}^{B^2}$ is the SH coefficients. In this paper, we only use the first $B=3$ bands of the SH basis to formulate the illumination model. To estimate the parameters $\bm{\chi}=\left\{ \bm{\alpha}_{id}, \bm{\alpha}_{exp}^j, \bm{\alpha}_{tex}, \bm{\tau}^j, \bm{\nu}^j \right\}$, $j=(1, 2, ..., N)$, we minimize the following objective function:
\begin{equation}
    \label{eq:proxy_objective}
    E(\bm{\chi}) = w_{im}E_{image} + w_{ld}E_{land} + w_{reg}E_{reg},
\end{equation}
where $E_{image}$ is the photometric consistency term, $E_{land}$ is the landmark term, and $E_{reg}$ is the regularization term. $w_{im}, w_{ld}$ and $w_{reg}$ are their corresponding weights and are empirically set as 50.0, 5.0, and 0.1, respectively. Specifically, the photometric consistency, aiming to minimize the difference between the observed intensity $\mathbf{I}_{in}^j \in \mathbb{R}^3$ and the rendered intensity $\mathbf{I}^j \in \mathbb{R}^3$ with Eq.~\ref{eq:image_formula}, is expressed as:
\begin{equation}
    E_{image}(\bm{\chi}) = \sum_{j=1}^N\displaystyle\frac{1}{|\mathcal{M}_h^j|}\left\| \mathbf{I}_{in}^j - \mathbf{I}^j \right\|_F^2,
\end{equation}
where $\mathcal{M}_h^j$ contains all pixels covered by the head model in the frame $j$, and $\|\cdot\|_F$ is the Frobenius norm. Since the facial structural information can be represented by facial landmarks, we introduce the landmark term to make the projections of 3D landmarks close to their corresponding landmarks on input frames:
\begin{equation}
    E_{land}(\bm{\chi}) = \sum_{j=1}^N\sum_{i \in \mathcal{L}} \left\| \mathbf{q}_i^j - \bm{\Pi}(\mathbf{R}^j\mathbf{V}_i^j + \mathbf{t}^j) \right\|_2^2,
\end{equation}
where $\mathcal{L}$ contains the landmark indexes. $\mathbf{q}_i^j$ is the 2D landmark location on the frame $j$ detected using \cite{bulat2017far}, and $\mathbf{V}_i^j$ is its corresponding 3D landmark location on the fitted head model. $\bm{\Pi}$, $\mathbf{R}^j$ and $\mathbf{t}^j$ are the perspective projection function, rotation matrix and translation vector in Eq.~\ref{eq:camera}. Finally, we utilize the regularization term to ensure that the parameters of the fitted parametric head model are plausible, which can be described as:
\begin{equation}
    E_{reg}(\bm{\chi}) = \left\| \bm{\alpha}_{id} \right\|_2^2 + \sum_{j=1}^N\left\| \bm{\alpha}_{exp}^j \right\|_2^2 + \left\| \bm{\alpha}_{tex} \right\|_2^2.
\end{equation}

\begin{figure}[htbp]
    \centering
    \includegraphics[width=0.48\textwidth]{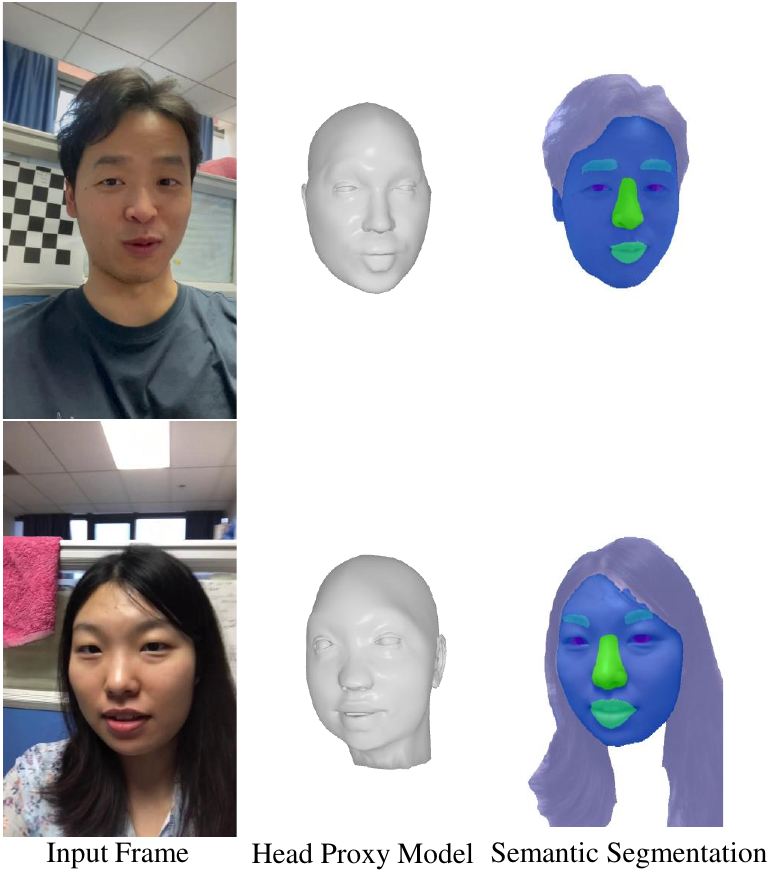}
    \caption{Two examples of reconstructed head proxy models and extracted semantic segmentation information. We show a single input frame from each video sequence on the left and its corresponding head proxy model and semantic segmentation information on the right.}
    \label{fig:proxylabel}
\end{figure}

When minimizing the objective function Eq.~\ref{eq:proxy_objective}, we assume that all the input frames share the same FLAME shape and texture parameters $\bm{\alpha}_{id}$, $\bm{\alpha}_{tex}$ while the FLAME expression parameters $\bm{\alpha}_{exp}$, the camera parameters $\bm{\tau}$ and the lighting parameters $\bm{\nu}$ are frame-dependent. We first initialize the FLAEM parameters and lighting parameters to zero, then let the Euler angles be zero to make rotation matrices unit matrices, and set the translation vectors as zero vectors. Then we solve this minimization problem using the gradient descent method until convergence. It takes about 2.8min to do the optimization with around 110 input frames on a single GeForce RTX 2080 Ti GPU. Fig.~\ref{fig:proxylabel} represents some reconstruction results of the above-mentioned method.

\subsection{Semantic Segmentation}
\label{sec:semantic}

To further improve the reconstruction accuracy and preserve the head structure, we introduce the semantic segmentation information to guide the reconstruction of the head model. We use the face parsing network in MaskGAN \cite{lee2020maskgan} to extract the head semantic segmentation information frame-by-frame. MaskGAN considers the task of semantic segmentation as an image-to-image translation problem and achieves the goal of face manipulation through semantic-based mapping. It adopts a similar network architecture to Pix2PixHD \cite{wang2018high}, and is trained on the datasets of CelebA-HQ \cite{karras2017progressive} and CelebAMask-HQ \cite{lee2020maskgan} at a resolution of 512 $\times$ 512 with paired segmentation masks. We use the pre-trained model to extract the head semantic segmentation information frame-by-frame, including the hair, face, eyebrows, eyes, mouse, and nose as shown in Fig.~\ref{fig:proxylabel}. The extraction time is approximately 0.21s of a single input frame on a single GeForce RTX 2080 Ti GPU. With this head semantic segmentation information, we perform a semantic neural network $s$ with trainable parameters $\bm{\zeta} \in \mathbb{R}^{n_s}$ to predict the probabilities of each segmentation part at a given point $\mathbf{x}_c \in \mathbb{R}^3$ in the canonical space with its global geometry feature vector $\mathbf{z}_{cg} \in \mathbb{R}^{256}$. We further design a loss term to guide the probability prediction as mentioned in Sec.~\ref{sec:loss}.

\subsection{Loss}
\label{sec:loss}

To reasonably utilize the self-supervision signals provided by input frames and constraint the implicit neural network $f$ to be a signed distance function, we first adopt two consistency loss terms and two regularization loss terms during the network training. Simultaneously, with the guidance of the above-mentioned head priors, we also put forward two prior-based loss terms into our framework to ensure the structural and depth information of the reconstructed head models while improving the reconstruction accuracy and robustness. In the following content, we first illustrate some symbols used in the loss term calculations and then describe all the loss terms in detail.

Our framework composes of five trainable network parameters, including $\bm{\beta}$ used in the offset estimation network $d$, $\bm{\gamma}$ used in the topology feature network $h$, $\bm{\eta}$ used in the SDF network $f$, $\bm{\theta}$ used in the rendering network $g$, $\bm{\zeta}$ used in the semantic prediction network $s$, and two trainable latent codes, including the latent rigid registration code $\mathbf{z}_r$ and the latent appearance code $\mathbf{z}_{cc}$. Meanwhile, the latent displacement code $\mathbf{z}_d$ is set as the expression parameter $\bm{\alpha}_{exp}$ obtained in Sec.~\ref{sec:proxy}. For a given pixel $p$ in the current frame $F$, let $\mathbf{v}_p$ be the viewing direction pointing from the camera center $\mathbf{c}_p$ towards $p$, and $R_p=\left\{ \mathbf{c}_p + t\mathbf{v}_p | t \geq 0 \right\}$ be the ray through pixel $p$. Then the sampled point $\mathbf{x}_p=\mathbf{x}_p(t)$ along the ray $R_p$ would be first rigidly registered from the current frame space into the canonical one as $\mathbf{x}_r = \mathbf{x}_r(t)$, then transformed to $\mathbf{x}_c = \mathbf{x}_c(t)$. With these variables, we can calculate the loss terms as follows.

\para{Photometric Consistency Term.} To reasonably utilize the self-supervision signals provided by input frames, we introduce the photometric consistency term to penalize the deviation between the observed intensity $\mathbf{I}_{in}$ in the current frame $F$ and the rendered appearance $\mathbf{\mathcal{C}}_{\bm{\beta, \gamma, \eta, \theta}}$ from Eq.~\ref{eq:rgb}, as \cite{yariv2020multiview}:
\begin{equation}
    \mathcal{L}_{rgb}(\bm{\beta}, \bm{\gamma}, \bm{\eta}, \bm{\theta}; \mathbf{z}_r, \mathbf{z}_{cc}) = \displaystyle\frac{1}{|\mathcal{M}|}\sum_{p \in \mathcal{M}} \left| \mathbf{I}_{in}(p) - \mathbf{\mathcal{C}}_{\bm{\beta, \gamma, \eta, \theta}}(p) \right|,
\end{equation}
where the input mask $\mathcal{M}$ of the current frame $F$ is the collection of all pixels in the 2D head region, obtained from the semantic segmentation result in Sec.~\ref{sec:semantic}, and $p \in \mathcal{M}$ is a pixel $p$ in the input mask.

\para{Contour Consistency Term.} Similar to \cite{wang2021neus}, we implement a mask term to ensure the contour consistency of the reconstructed 3D head model and the 2D input mask of the current frame $F$, which is formulated as:
\begin{equation}
    \mathcal{L}_{mask}(\bm{\beta}, \bm{\gamma}, \bm{\eta}; \mathbf{z}_r) = \displaystyle\frac{1}{\left| \mathcal{P} \right|}\sum_{p \in \mathcal{P}} BCE\left( O_p, \hat{O}_w(p) \right),
\end{equation}
where $\mathcal{P}$ is the set of all sampling pixels. $O_p$ is an indicator function identifying whether pixel $p$ is in the input mask $\mathcal{M}$, and $\hat{O}_w$ is the sum of weights along the ray $R_p$ indicating whether $p$ is occupied by the rendered object:
\begin{equation}
	\hat{O}_w(p) = \int_{t_n}^{t_f}T_{sm}(t)\sigma(\mathbf{x}_c(t))dt.
\end{equation}

\para{Latent Code Regularization Term.} The latent code regularization term aims to make the two latent codes of the current frame $F$ compact in the latent spaces:
\begin{equation}
    \mathcal{L}_{creg}(\mathbf{z}_r, \mathbf{z}_{cc}) = \| \mathbf{z}_r \|_2^2 + \| \mathbf{z}_{cc} \|_2^2.
\end{equation}

\para{Eikonal Regularization Term.} The Eikonal regularization term \cite{gropp2020implicit} constraints the network $f$ to be approximately a singed distance function in both the current frame space and the canonical one:
\begin{equation}
    \begin{aligned}
    \mathcal{L}_{eik}(\bm{\beta}, \bm{\gamma}, \bm{\eta}; \mathbf{z}_r) &= \mathbb{E}_{\mathbf{x}_p}\left( \| \nabla_{\mathbf{x}_p}f(\mathbf{x}_c, \mathbf{w}_h | \bm{\eta})\| - 1 \right)^2 \\
    &+ \mathbb{E}_{\mathbf{x}_c}\left( \| \nabla_{\mathbf{x}_c}f(\mathbf{x}_c, \mathbf{w}_h | \bm{\eta}) \| - 1  \right)^2.
    \end{aligned}
\end{equation}

\para{Head Prior Geometry Term.} Based on the head proxy models obtained in Sec.~\ref{sec:proxy}, we formulate a head prior geometry term to provide the initial geometric information and depth range for the implicit neural head model, which can be described as:
\begin{equation}
\label{eq:loss_proxy}
    \mathcal{L}_{pg}(\bm{\beta}, \bm{\gamma}, \bm{\eta}; \mathbf{z}_r) = \displaystyle\frac{1}{|\mathcal{V}|}\sum_{\mathbf{V} \in \mathcal{V}}\left| f\left( \mathbf{V}_c, \mathbf{w}_h^v \ | \ \bm{\eta} \right) \right|,
\end{equation}
where $\mathcal{V}$ contains the sampling points on the head proxy model fitted from the current frame $F$. $\mathbf{V}_c \in \mathbb{R}^3$ is the corresponding point of $\mathbf{V} \in \mathbb{R}^3$ after non-rigid deformations described in Sec.~\ref{sec:ddf}, and $\mathbf{w}_h^v$ is the ambient coordinate obtained using the topology feature network $h$.

\para{Head Semantic Term.} With the semantic segmentation information in Sec.~\ref{sec:semantic}, the head semantic term is introduced to further maintain the geometric accuracy:
\begin{equation}
\label{eq:loss_semantic}
    \mathcal{L}_{se}(\bm{\beta}, \bm{\gamma}, \bm{\eta}, \bm{\zeta}; \mathbf{z}_r) = \displaystyle\frac{1}{|\mathcal{M}|}\sum_{p \in \mathcal{M}} CE(S(p), L_p),
\end{equation}
where $CE$ is the cross-entropy loss. $L_p$ is the extracted 2D semantic segmentation information in Sec.~\ref{sec:semantic}, and $S(p)$ indicates the predicted semantic segmentation probability of the ray $R_p$ computed with:
\begin{equation}
    S(p) = \int_{t_n}^{t_f}T_{sm}(t)\sigma(\mathbf{x}_c(t))s(\mathbf{x}_c(t), \mathbf{z}_{cg} \ | \ \bm{\zeta})dt,
\end{equation}
where $\mathbf{z}_{cg}$ is the global geometry feature vector as described in Sec.~\ref{sec:implement}, and $s(\mathbf{x}_c(t), \mathbf{z}_{cg} \ | \ \bm{\zeta})$ is the predicted semantic segmentation probability of the point $\mathbf{x}_c$.

Combining these loss terms, our loss function could be written as:
\begin{equation}
\label{eq:final_loss}
    \begin{aligned}
    & \quad \ \mathcal{L}(\bm{\beta}, \bm{\gamma}, \bm{\eta}, \bm{\theta}, \bm{\zeta}; \mathbf{z}_r, \mathbf{z}_{cc}) \\
    &= w_{rgb}\mathcal{L}_{rgb}(\bm{\beta}, \bm{\gamma}, \bm{\eta}, \bm{\theta}; \mathbf{z}_r, \mathbf{z}_{cc}) + w_{cr}\mathcal{L}_{creg}(\mathbf{z}_r, \mathbf{z}_{cc}) \\
    &+ w_{m}\mathcal{L}_{mask}(\bm{\beta}, \bm{\gamma}, \bm{\eta}; \mathbf{z}_r)  + w_{eik}\mathcal{L}_{eik}(\bm{\beta}, \bm{\gamma}, \bm{\eta}; \mathbf{z}_r) \\
    &+ w_{p}\mathcal{L}_{pg}(\bm{\beta}, \bm{\gamma}, \bm{\eta}; \mathbf{z}_r) + w_{s}\mathcal{L}_{se}(\bm{\beta}, \bm{\gamma}, \bm{\eta}, \bm{\zeta}; \mathbf{z}_r).
    \end{aligned}
\end{equation}
With the well-designed loss function, our method can reconstruct high-fidelity complete 3D head models frame-by-frame extracted from the signed distance fields using the Marching Cubes algorithm \cite{lorensen1987marching}.

% \begin{equation}
%    \begin{aligned}
%    \mathcal{L}_{eik}(\bm{\beta}, \bm{\gamma}, \bm{\eta}; \mathbf{z}_r, \mathbf{z}_d) &= \mathbb{E}_{x_p}\left( \| % f(\mathbf{x}_p, \mathbf{w}_h \ | \ \bm{\eta}) - 1 \| \right)^2 \\
%   &+ \mathbb{E}_{x_c}\left( \| f(\mathbf{x}_c, \mathbf{w}_h \ | \ \bm{\eta}) - 1 \| \right)^2.
%    \end{aligned}
% \end{equation}

% The latent code regularization term aims to make the three latent codes of the current frame $F$ obey the normal distribution $\mathcal{N}(0, 1)$:
\section{Experiments}

\subsection{Implementation Details}
\label{sec:implement}

\textbf{Evaluation Data.} We mainly evaluate our approach on three kinds of datasets with real monocular dynamic video sequences. Firstly, we collected 4 real monocular dynamic video sequences with a mobile phone ranging from 0 to 180 degrees. Then, we extracted an average of 110 frames under different views from each video sequence at a resolution of 1920 $\times$ 1080. To quantitatively evaluate the reconstruction accuracy of our method, we also captured 4 RGB-D monocular dynamic video sequences with an iPhone X camera at a resolution of 640 $\times$ 480. We take around 110 RGB frames of each video sequence as input to our framework, and obtain their corresponding point clouds from the depth maps for geometric accuracy calculations. Additionally, to further verify the effectiveness of our method, we conduct tests on two publicly available monocular dynamic video sequences respectively obtained from NerFace \cite{gafni2021dynamic} and YouTube \cite{youtube}.

\textbf{Network Architecture.} Our framework is composed of five MLP networks. The offset estimation network $d(\mathbf{x}, \mathbf{z}_d \ | \ \bm{\beta})$ takes a 3D point $\mathbf{x}$ and the frame-specific latent displacement code $\mathbf{z}_d$ as input, and outputs a 3-dimensional offset. $d$ consists of 7 layers with hidden layers of width 128 and a skip connection from the input to the middle layer. The topology feature network $h(\mathbf{x}, \mathbf{z}_d \ | \ \bm{\gamma})$ comprises 7 layers with hidden layers of width 64 and a skip connection from the input to the middle layer, estimating a 2-dimensional ambient coordinate for topology features at $\mathbf{x}$ with $\mathbf{z}_d$. The implicit neural representation network $F_s(\mathbf{x}, \mathbf{w}_h \ | \ \bm{\eta}) = \left( f(\mathbf{x}, \mathbf{w}_h \ | \ \bm{\eta}), \mathbf{z}_{cg}(\mathbf{x}, \mathbf{w}_h \ | \ \bm{\eta}) \right)$ outputs a 1-dimensional signed distance value $d_s = f(\mathbf{x}, \mathbf{w}_h \ | \ \bm{\eta})$ and a 256-dimensional global geometry feature vector $\mathbf{z}_{cg} = \mathbf{z}_{cg}(\mathbf{x}, \mathbf{w}_h \ | \ \bm{\eta})$ of a given point $\mathbf{x}$ and its corresponding ambient coordinate $\mathbf{w}_h$. $F_s$ incorporates 8 layers with hidden layers of width 512 and a skip connection from the input to the middle layer. The rendering network $g(\mathbf{x}, \mathbf{n}, \mathbf{v}, \mathbf{z}_{cg}, \mathbf{z}_{cc} \ | \ \bm{\theta})$ has 4 layers with hidden layers of width 512, rendering the RGB value at $\mathbf{x}$ with its normal vector $\mathbf{n}$ and global geometry feature vector $\mathbf{z}_{cg}$ given the viewing direction $\mathbf{v}$ and the per-frame latent appearance code $\mathbf{z}_{cc}$. In these networks, positional encoding \cite{tancik2020fourier} is applied to the point $\mathbf{x}$ with 6 frequencies, the ambient coordinate $\mathbf{w}_h$ with 1 frequency and the viewing direction $\mathbf{v}$ with 4 frequencies. Meanwhile, the coarse-to-fine annealing technique is implemented to the positional encoding adopted from \cite{park2021nerfies}. In addition, the semantic prediction network $s(\mathbf{x}, \mathbf{z}_{cg} \ | \ \bm{\zeta})$ contains 4 layers with 512-dimensional hidden layers, predicting the semantic probabilities of the given point $\mathbf{x}$ with $\mathbf{z}_{cg}$.

\textbf{Optimization.} Our prior-guided dynamic implicit neural network requires optimizations for every subject, that takes as input a monocular dynamic video sequence with its per-frame corresponding masks and camera parameters. In this paper, we obtain the masks from the semantic segmentation information as mentioned in Sec.~\ref{sec:semantic}, and the camera parameters using the optimization-based method in Sec.~\ref{sec:proxy}. At each optimization epoch, We first input the randomly sampled points from the head proxy model recovered in Sec.~\ref{sec:proxy} for each input frame to the implicit neural representation network $F_s$ after non-rigid deformations described in Sec.~\ref{sec:ddf}. At the same time, we randomly sample some camera rays from all pixels in each input frame and obtain some sampled points $\mathbf{x} = \mathbf{c} + t\mathbf{v}$ along the rays using the coarse-to-fine sampling strategy in \cite{mildenhall2020nerf}. Then we transform these points to $\mathbf{x}_c$ in the canonical space using the proposed two-part dynamic deformation fields, and modify the viewing direction $\mathbf{v}$ as follows:
\begin{equation}
    \mathbf{v}_c = \mathbf{J}_{\mathbf{x}_c}(\mathbf{x}) \cdot \mathbf{v},
\end{equation}
where $\mathbf{J}_{\mathbf{x}_c}(\mathbf{x}) \in \mathbb{R}^{3 \times 3}$ is the Jacobian matrix of $\mathbf{x}_c$ with respect to $\mathbf{x}$.

In our experiments, we set a batch size of 256 sampling camera rays in all input frames simultaneously. We further sample 64 points along the rays in the coarse sampling stage, and additional 64 points in the fine sampling stage. We empirically set $w_{rgb}$, $w_{cr}$, $w_{m}$, $w_{eik}$, $w_{p}$ and $w_{s}$ as 1.0, 0.1, 0.1, 0.3, 0.8 and 0.1, respectively. We implement our approach in PyTorch \cite{paszke2017automatic} and optimize the network parameters and latent codes using the Adam solver \cite{kingma2014adam} for 2000 epochs. It takes about 13.75h to do the optimizations on a single GeForce RTX 2080 Ti GPU with around 110 input frames.

\begin{figure}[htbp]
    \centering
    \includegraphics[width=0.50\textwidth]{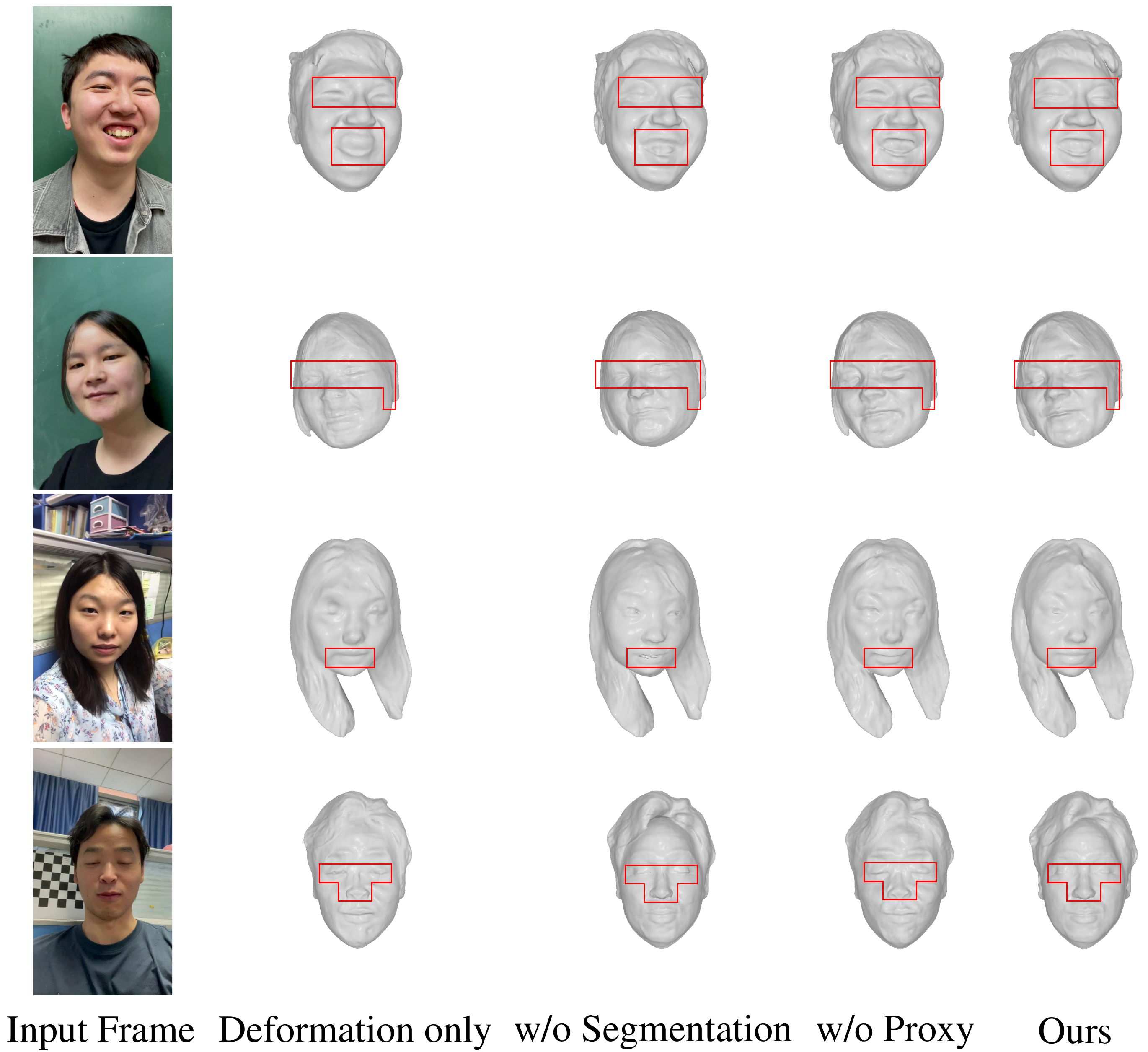}
    \caption{Ablation studies that compare the proposed method with three approaches that exclude all the head prior guidance, the head semantic segmentation information, and the head prior geometry respectively. For each method, we show the reconstructed 3D head models and a single input frame from each video sequence.}
    \label{fig:ablation}
\end{figure}

% Ablation studies that compare the proposed method with three approaches that respectively exclude all the head prior guidance (only using the dynamic deformation fields), the head semantic segmentation information, and the head prior geometry. On the left, we show a single input frame from each video sequence. On the right, we show their corresponding reconstructed 3D head models of different settings.

\subsection{Ablation Study} 

To verify the effectiveness of the prior guidance, we compare the proposed approach with alternative strategies by excluding different components. Firstly, we conduct an experiment on our baseline framework by eliminating both the head prior geometry loss term Eq.~\ref{eq:loss_proxy} and the head semantic loss term Eq.~\ref{eq:loss_semantic} from the loss function Eq.~\ref{eq:final_loss}. Secondly, the head semantic loss term Eq.~\ref{eq:loss_semantic} is excluded to show the influence of the head semantic segmentation information. Finally, we perform another experiment to demonstrate the capability of the head prior geometry by excluding the head prior geometry loss term Eq.~\ref{eq:loss_proxy}. Both the qualitative and quantitative comparison results of some monocular dynamic video sequences are shown in Fig.~\ref{fig:ablation} and Tab.~\ref{tab:ablation} respectively. It can be observed that the exclusion of any component would cause a degradation in the performance. In particular, the head prior geometry and the semantic segmentation information both contribute to the improvement of the reconstruction accuracy. Moreover, the head prior geometry provides an initial geometric structure and depth range for further implicit reconstructions.

%To verify the effectiveness of the dynamic deformation fields and the prior guidance, we compare the proposed approach with alternative strategies by excluding different components. Firstly, we conduct an experiment on the baseline method (NeuS \cite{wang2021neus}) without using the dynamic deformation fields and any prior loss term. Secondly, we illustrate the necessity of the dynamic deformation fields by conducting an experiment that eliminates both the head prior geometry loss term Eq.~\ref{eq:loss_proxy} and the head semantic loss term Eq.~\ref{eq:loss_semantic} from the loss function Eq.~\ref{eq:final_loss}. Thirdly, the head semantic loss term Eq.~\ref{eq:loss_semantic} is excluded to show the influence of the head semantic segmentation information. Finally, we perform another experiment to demonstrate the capability of the head prior geometry by excluding the head prior geometry loss term Eq.~\ref{eq:loss_proxy}. Both the qualitative and quantitative comparison results of some monocular dynamic video sequences are shown in \hl{Fig.} and \hl{Tab.} respectively. It can be observed that the exclusion of any component would cause a degradation in the performance. In particular, the dynamic deformation fields could describe the pose and expression changes of the captured speaker. The head prior geometry and the semantic segmentation information both contribute to the improvement of the reconstruction accuracy. Moreover, the head prior geometry provides an initial geometric structures and depth range for further implicit reconstructions.

\begin{table}[h]
	\centering  
	\caption{Average geometric errors (mm) on the RGB-D monocular dynamic video sequences of the ablation study. We compute the Charmfer distance of the target head regions on the point clouds with our reconstructed head models.}
	\label{tab:ablation}
	\centering
	\begin{tabular}{ccccc}  
		\toprule   
		\# Input & Deformation only & w/o Segmentation & w/o Proxy & Ours \\  
		\midrule   
		110 & 0.0477 & 0.0424 & 0.0410 & \textbf{0.0395} \\  
		\bottomrule  
	\end{tabular}
\end{table}

% Average geometric errors (mm) on the RGB-D monocular dynamic video sequences of the ablation study. We take around 110 frames for each video sequence as input and compute the Charmfer distance of the target head regions on the point clouds, generated from their corresponding depth maps, with our reconstructed head models.

\subsection{Comparisons}

\begin{figure}[htbp]
    \centering
    \includegraphics[width=0.50\textwidth]{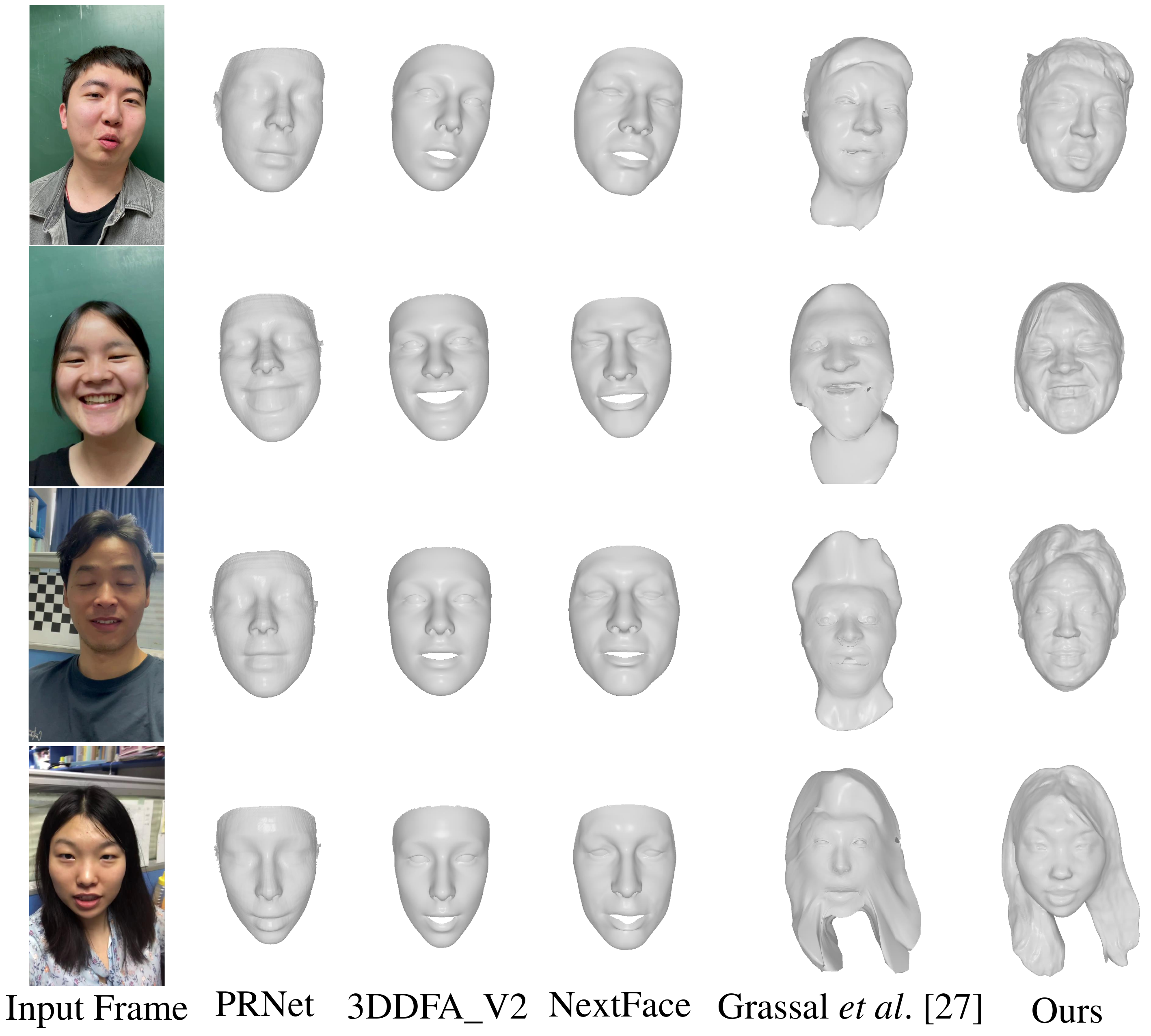}
    \caption{Qualitative comparisons of PRNet \cite{feng2018prn}, 3DDFA\_V2 \cite{guo2020towards,3ddfa_cleardusk}, NextFace \cite{dib2021practical,dib2021towards,dib2022s2f2}, Grassal \etal~\cite{grassal2022neural} and our method on the phone-captured RGB video sequences. For each method, we show the reconstructed model for a single input frame.}
    \label{fig:phone_quality}
\end{figure}

% Qualitative comparisons of PRNet \cite{feng2018prn}, 3DDFA\_V2 \cite{guo2020towards,3ddfa_cleardusk}, NextFace \cite{dib2021practical,dib2021towards,dib2022s2f2}, Grassal \etal~\cite{grassal2022neural} and our method on the phone-captured RGB video sequences. The single input frame of each video is shown in the left, and the reconstructed models of different methods are shown in the right.

\begin{figure}[htbp]
    \centering
    \includegraphics[width=0.50\textwidth]{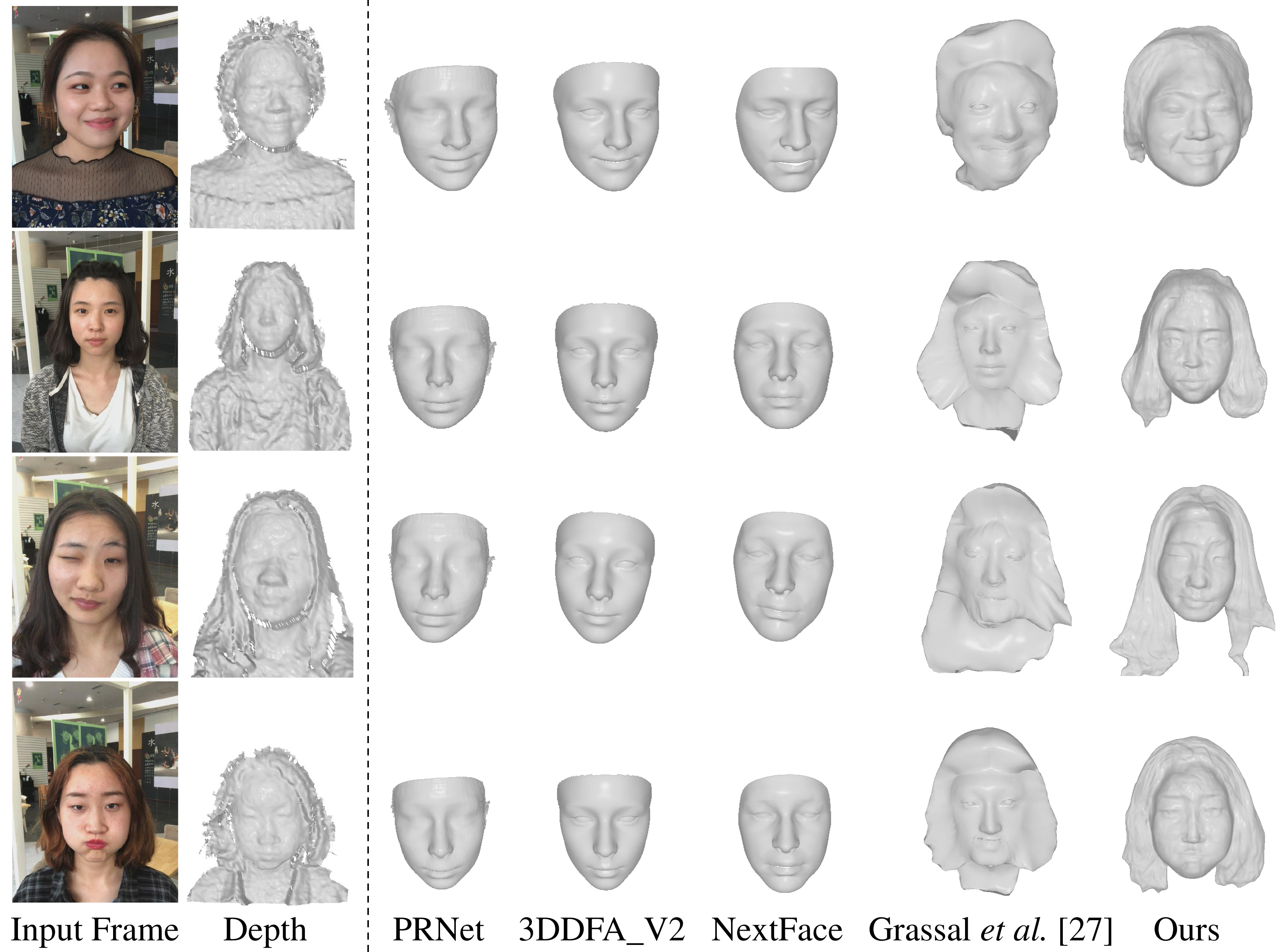}
    \caption{Reconstruction results of PRNet \cite{feng2018prn}, 3DDFA\_V2 \cite{guo2020towards,3ddfa_cleardusk}, NextFace \cite{dib2021practical,dib2021towards,dib2022s2f2}, Grassal \etal~\cite{grassal2022neural} and our method on the iPhone X-captured RGB-D video sequences. We show the reconstructed models of all the five methods.}
    \label{fig:phone_rgbd}
\end{figure}

\textbf{Comparisons with 3D face reconstructions from monocular videos.} To evaluate the geometric accuracy of the face region of our reconstructed head model, we mainly compare our prior-guided dynamic implicit volumetric rendering method with PRNet \cite{feng2018prn}, 3DDFA\_V2 \cite{guo2020towards,3ddfa_cleardusk} and NextFace \cite{dib2021practical,dib2021towards,dib2022s2f2}. Both PRNet \cite{feng2018prn} and 3DDFA\_V2 \cite{guo2020towards,3ddfa_cleardusk} are deep learning-based methods. PRNet \cite{feng2018prn} designs a 2D UV position map to represent a 3D face model and regresses it via a simple convolutional neural network. Based on a lightweight backbone network, 3DDFA\_V2 \cite{guo2020towards,3ddfa_cleardusk} proposes a meta-joint optimization strategy to regress 3DMM parameters. While NextFace \cite{dib2021practical,dib2021towards,dib2022s2f2} adopts a three-step optimization method to gradually regress and refine 3DMM parameters. All the three methods are based on low-dimensional explicit representations, thus they could only recover low-frequency facial geometric structures. Firstly, we compare our method with these three methods on our phone-captured RGB and RGB-D monocular dynamic video sequences. For all the methods, we take approximately 110 RGB frames for each video sequence as input and compute the geometric accuracy of the reconstructed face region with the target face region on the point cloud frame-by-frame, generated from the corresponding depth map. The qualitative comparison results shown in Fig.~\ref{fig:phone_quality} demonstrate that our method can better describe the identity and expression information of the face region. For quantitative comparisons, we calculate the geometric error of the reconstructed model and the point cloud in the target face region for each input frame, by firstly applying a transformation with seven degrees of freedom (six for rigid transformation and one for scaling) to align their face regions and then computing the Charmfer distance between them. Some reconstruction results of the RGB-D monocular dynamic video sequences are shown in Fig.~\ref{fig:phone_rgbd}, and the average geometric errors of PRNet \cite{feng2018prn}, 3DDFA\_V2 \cite{guo2020towards,3ddfa_cleardusk}, NextFace \cite{dib2021practical,dib2021towards,dib2022s2f2} and ours are shown in Tab.~\ref{tab:quantity-rgbd}.
% of the target face regions on the ground truth models

\begin{figure}[htbp]
    \centering
    \includegraphics[width=0.50\textwidth]{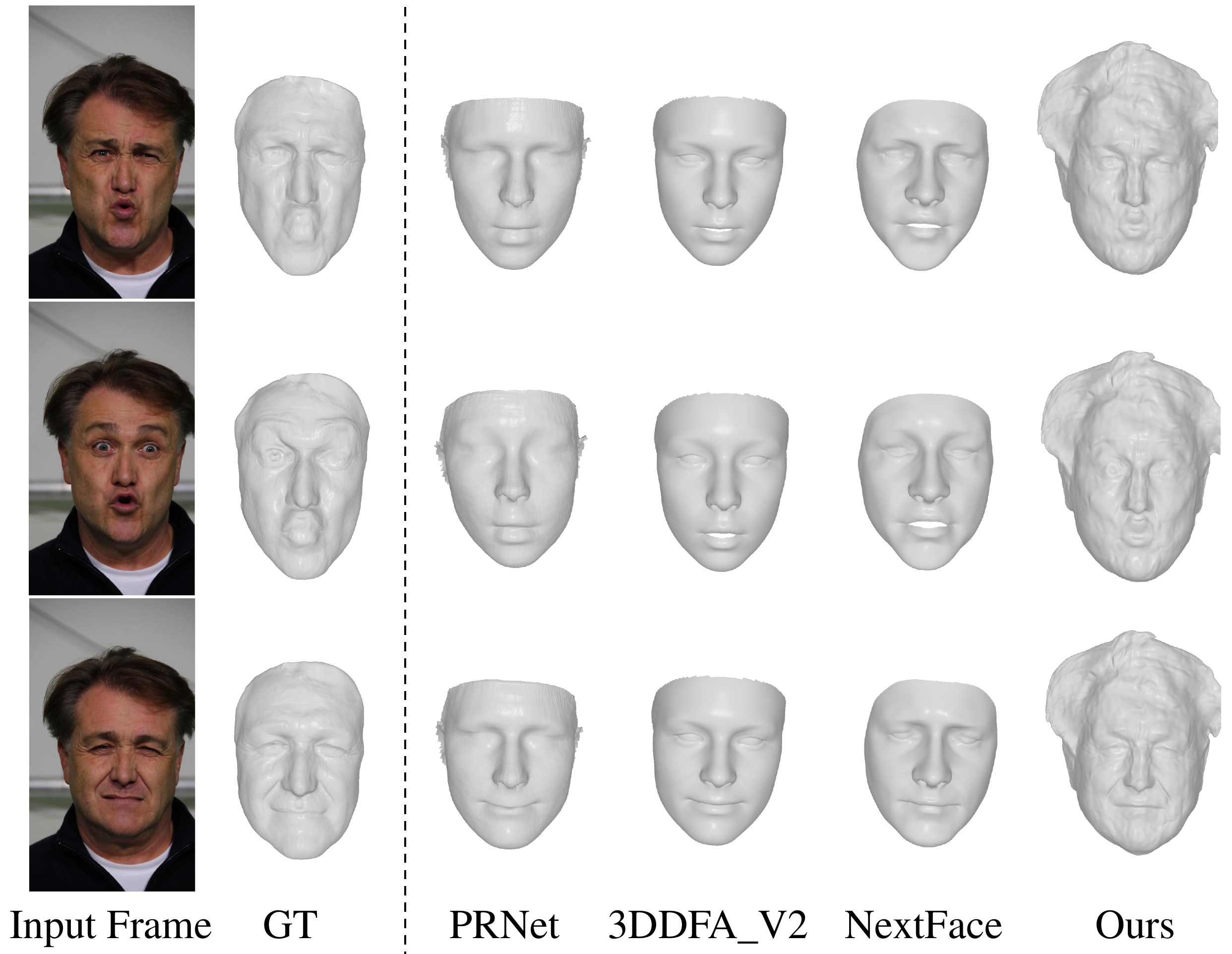}
    \caption{Reconstructed models of PRNet \cite{feng2018prn}, 3DDFA\_V2 \cite{guo2020towards,3ddfa_cleardusk}, NextFace \cite{dib2021practical,dib2021towards,dib2022s2f2} and our method on the video sequence provided by \cite{valgaerts2012lightweight}. We show three reconstruction results of three input frames for each method.}
    \label{fig:facecap}
\end{figure}

Then, we additionally perform a comparison on the video sequence provided by \cite{valgaerts2012lightweight}. In this case, we take 100 frames as input to all of the methods and compute the geometric errors as the Charmfer distances of the face regions between the reconstructed models and the ground truth models in \cite{valgaerts2012lightweight} frame-by-frame. The average reconstruction errors of PRNet \cite{feng2018prn}, 3DDFA\_V2 \cite{guo2020towards,3ddfa_cleardusk}, NextFace \cite{dib2021practical,dib2021towards,dib2022s2f2} and our method are shown in Tab.~\ref{tab:quantity-facecap}, and three reconstruction examples are shown in Fig.~\ref{fig:facecap}. These two comparison experiments indicate that our approach can effectively improve the facial reconstruction accuracy.

% We show three input frames in the first column, and their corresponding ground truth models in the second column. In the right, we show reconstruction results of all the four methods.

\begin{table}[h]
	\centering  
	\caption{Quantitative comparisons of PRNet \cite{feng2018prn}, 3DDFA\_V2 \cite{guo2020towards,3ddfa_cleardusk}, NextFace \cite{dib2021practical,dib2021towards,dib2022s2f2} and our method on the video sequence in \cite{valgaerts2012lightweight}. The average geometric error (mm) is shown below, computed as the Charmfer distance of the target face regions on the ground truth models with the face regions on our reconstructed models.}
	\label{tab:quantity-facecap}
	\centering
	\scalebox{0.835}
	{
	\begin{tabular}{ccccc}  
		\toprule   
		\# Input & PRNet \cite{feng2018prn} & 3DDFA\_V2 \cite{guo2020towards,3ddfa_cleardusk} & NextFace \cite{dib2021practical,dib2021towards,dib2022s2f2} & Ours \\  
		\midrule   
		100 & 6.0244 & 6.7099 & 6.7637 & \textbf{4.5940} \\  
		\bottomrule  
	\end{tabular}
	}
\end{table}
% We take 100 frames as input and compute the Charmfer distance of the target face regions on the ground truth models, provided by \cite{valgaerts2012lightweight}, with the face regions on our reconstructed models.

% A single input frame of each video sequence is shown in the first column. Their corresponding depth information is shown in the second column. And the reconstructed models of all the five methods are shown in the right.

\begin{table}[h]
	\centering  
	\caption{Average geometric errors (mm) on the RGB-D video sequences of PRNet \cite{feng2018prn}, 3DDFA\_V2 \cite{guo2020towards,3ddfa_cleardusk}, NextFace \cite{dib2021practical,dib2021towards,dib2022s2f2} and our method. We compute the Charmfer distance of the target face regions on the point clouds with the face regions on our reconstructed head models.}
	\label{tab:quantity-rgbd}
	\centering
	\scalebox{0.835}{
	\begin{tabular}{cccccc}  
		\toprule   
		\# Input & PRNet \cite{feng2018prn} & 3DDFA\_V2 \cite{guo2020towards,3ddfa_cleardusk} & NextFace \cite{dib2021practical,dib2021towards,dib2022s2f2} & ours \\  
		\midrule   
		110 & 0.0266 & 0.0276 & 0.0309 & \textbf{0.0196} \\  
		\bottomrule  
	\end{tabular}}
\end{table}
% We take around 110 frames for each video sequence as input and compute the Charmfer distance of the target face regions on the point clouds, generated from their corresponding depth maps, with the face regions on our reconstructed head models.

\begin{figure}[htbp]
    \centering
    \includegraphics[width=0.50\textwidth]{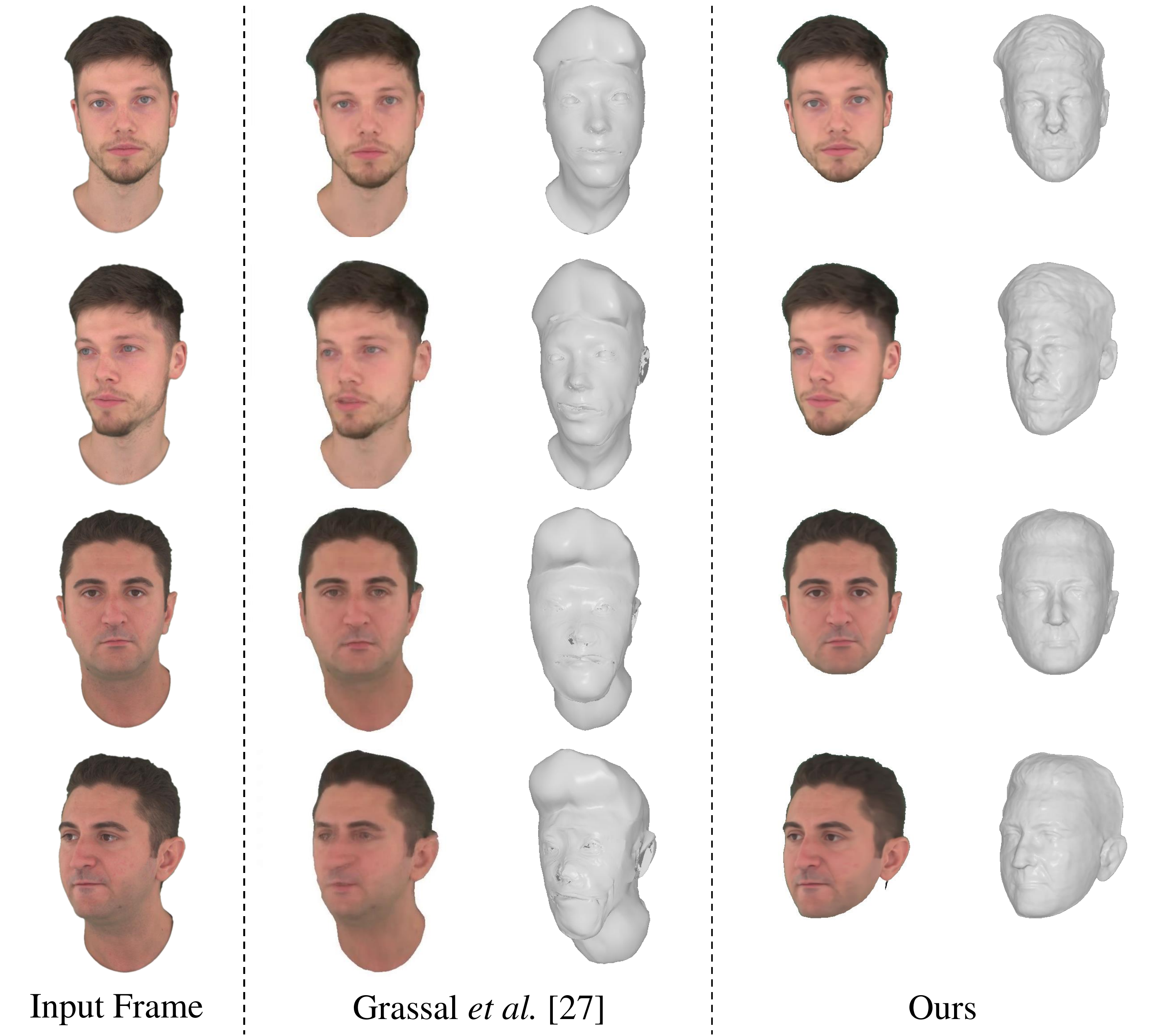}
    \caption{Qualitative comparisons with Grassal \etal~\cite{grassal2022neural} and our method on two monocular dynamic sequences provided by \cite{grassal2022neural}. We show two input frames for each video sequence in the left, And the corresponding rendered results and reconstruction models are shown in the right.}
    \label{fig:quality_neural}
\end{figure}

\textbf{Comparisons with 3D head reconstructions from monocular videos.} We compare our method with a state-of-the-art monocular video-based 3D head reconstruction method proposed by Grassal \etal~\cite{grassal2022neural}. This method first fits a FLAME head model from each input frame and then computes the point-wise offsets to further recover the hair region and refine the face region. It performs well on video sequences where both sides of the head are visible at least once. Different from it, we only utilize the FLAME head model as a piece of prior knowledge to improve the reconstruction accuracy and robustness, and employ an SDF-based implicit representation to describe the 3D head models. For qualitative comparisons, we compare the two methods on our phone-captured RGB monocular dynamic video sequences and two video sequences provided by Grassal \etal~\cite{grassal2022neural}. For a fair comparison, we take around 110 frames of our phone-captured monocular dynamic video sequence as input to the two methods and adjust the optimization epochs of Grassal \etal~\cite{grassal2022neural} to 1750 according to their descriptions. On the two video sequences provided by them, we approximately input 120 frames for each video sequence to the two methods and optimize their method for a total of 1500 epochs. Qualitative comparison results shown in Fig.~\ref{fig:phone_quality} and Fig.~\ref{fig:quality_neural} illustrate that our method can recover the 3D head models more robustly and accurately. As for quantitative comparisons, we carry out experiments on our RGB-D monocular dynamic video sequences captured by an iPhone X camera. We take about 110 RGB frames of each video sequence as input to the two methods. Then we compute the geometric errors frame-by-frame by firstly aligning the reconstructed 3D head model with the target head region on the point cloud generated from the corresponding depth map and calculating the Charmfer distance between them. Some comparison examples are shown in Fig.~\ref{fig:phone_rgbd} and the average geometric errors of Grassal \etal~\cite{grassal2022neural} and ours are 0.0569mm and \textbf{0.0395mm} respectively. Both the two comparison experiments demonstrate that our method outperforms the other approach due to the prior guidance and the flexible dynamic implicit neural representation.

\begin{figure*}[htbp]
    \centering
    \includegraphics[width=1.0\textwidth]{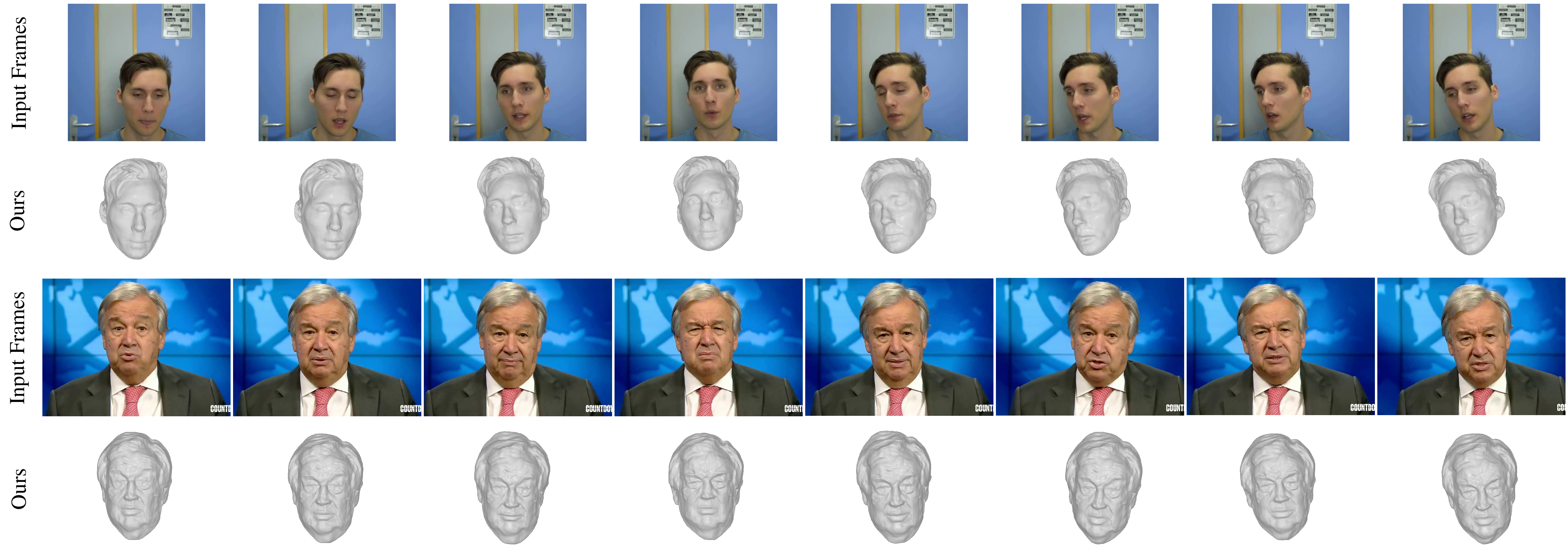}
    \caption{Evaluations on two publicly available monocular dynamic video sequences obtained from NerFace \cite{gafni2021dynamic} and YouTube \cite{youtube} respectively. We show eight example input frames and their corresponding reconstructed 3D head models for each input video sequence.}
    \label{fig:quality_addtion}
\end{figure*}

\subsection{Additional Evaluations}

To further validate the effectiveness of our method on different kinds of monocular dynamic video sequences, we additionally carry out two experiments on two publicly available monocular dynamic video sequences respectively obtained from NerFace \cite{gafni2021dynamic} and YouTube \cite{youtube}. For the monocular dynamic video sequence provided by NerFace \cite{gafni2021dynamic}, we extracted 107 frames at a resolution of 512 $\times$ 512 as input to our method. Meanwhile, we extracted 120 frames at a resolution of 1280 $\times$ 720 from the monocular dynamic video sequence collected from YouTube \cite{youtube}. Some reconstruction results are shown in Fig.~\ref{fig:quality_addtion}. It can be seen that our proposed approach can achieve robust and high-fidelity reconstruction results due to the utilization of our dynamic implicit volumetric rendering framework with the head prior guidance.

% We select around 120 frames as input for each video sequence. We show two input frames for each video sequence in the left, And the corresponding rendered results and reconstruction models are shown in the right.

\subsection{Application: Facial Reenactment}

\begin{figure}[htbp]
    \centering
    \includegraphics[width=0.45\textwidth]{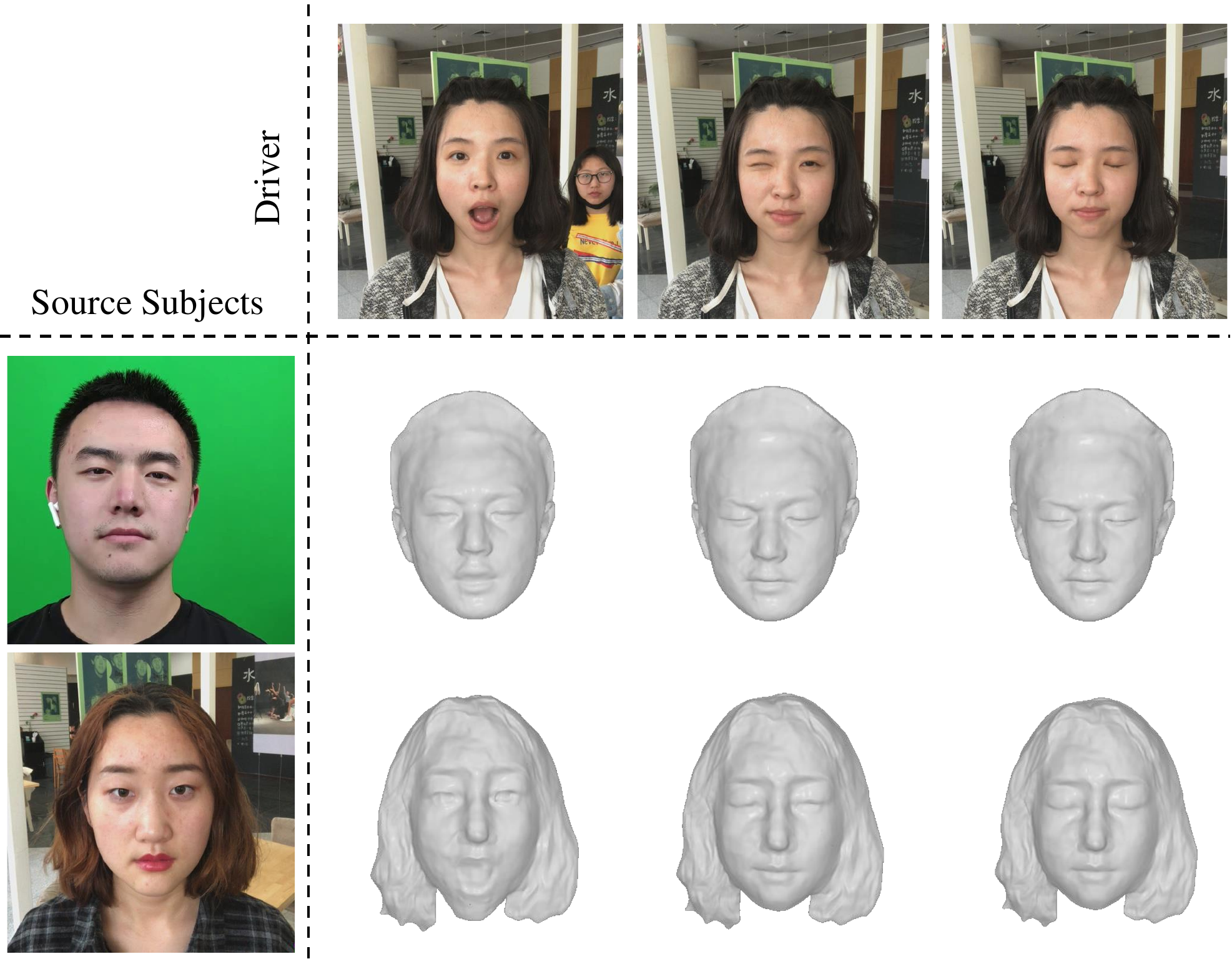}
    \caption{Some reenactment results between different subjects. Three frames of the driving sequence are shown on the top. Two source subjects are shown in the left and their corresponding reenactment models are shown in the middle.}
    \label{fig:application}
\end{figure}

Benefit from the proposed dynamic deformation fields, we can achieve the reenactment of a pre-reconstructed 3D head sequence according to a driving sequence. In this experiment, we first obtain the pose information $\bm{\tau}$ and the expression parameter $\bm{\alpha}_{exp}$ of the driving sequence frame-by-frame as described in Sec.~\ref{sec:proxy}. Then we use $\bm{\alpha}_{exp}$ to estimate the point-wise offset and topology changes at the given pose $\bm{\tau}$. Finally, we can extract the reenacted head models from the pre-optimized neural network $f$. Some reenactment results are shown in Fig.~\ref{fig:application}. It can be observed that our method can realize the reenactment between different subjects, which indicates the potentiality of our method in the digital human head-related applications.

\section{Conclusion}
We proposed a prior-guided dynamic implicit volumetric rendering network to reconstruct a high-fidelity 3D head model frame-by-frame from a monocular dynamic video sequence. We first conduct a two-part dynamic deformation field to describe the head motion and deformation, and then we extract two main head priors from the input video frames, including the head prior geometry and semantic segmentation information. With the guidance of this prior knowledge, we design two prior-based loss terms for our dynamic implicit volumetric rendering network to resolve the depth ambiguity while improving the reconstruction robustness and accuracy. Extensive experiments demonstrated that our method outperforms existing face and head reconstruction methods from monocular video sequences. Additional evaluations illustrated that our method performs well on various types of monocular dynamic video sequences. Furthermore, with our novel algorithm design, the reconstruction results were directly used for facial reenactment, which verifies the potentiality of our method in the digital human-related applications.

% In this paper, we propose a prior-guided dynamic implicit volumetric rendering network to reconstruct a high-fidelity integrated 3D head model frame-by-frame from a monocular dynamic video sequence. We first conduct a two-part dynamic deformation field to describe the head motion, and then we extract two main head priors from the input video frames, including the head prior geometry and semantic segmentation information. In particular, the head prior geometry provides an initial head geometric structure and a reasonable depth range, while the semantic segmentation information further maintains the geometric accuracy. Thus, with the guidance of this prior knowledge, we design two prior-based loss terms for our dynamic implicit volumetric rendering network to avoid the depth ambiguity and improve the reconstruction robustness and accuracy. Extensive experiments illustrate that our method outperforms some state-of-the-art face and head reconstruction methods from monocular video sequences. Additional evaluations demonstrate that our method performs well on various types of monocular dynamic video sequences.

\noindent \textbf{Acknowledgement} 
This research was supported by the National Natural Science Foundation of China (No.62122071) and the Fundamental Research Funds for the Central Universities (No.WK3470000021).

\normalem
{\small
	\bibliographystyle{IEEEtran}
	\bibliography{mybibfile}

\begin{thebibliography}{10}
\providecommand{\url}[1]{#1}
\csname url@samestyle\endcsname
\providecommand{\newblock}{\relax}
\providecommand{\bibinfo}[2]{#2}
\providecommand{\BIBentrySTDinterwordspacing}{\spaceskip=0pt\relax}
\providecommand{\BIBentryALTinterwordstretchfactor}{4}
\providecommand{\BIBentryALTinterwordspacing}{\spaceskip=\fontdimen2\font plus
\BIBentryALTinterwordstretchfactor\fontdimen3\font minus
  \fontdimen4\font\relax}
\providecommand{\BIBforeignlanguage}[2]{{%
\expandafter\ifx\csname l@#1\endcsname\relax
\typeout{** WARNING: IEEEtran.bst: No hyphenation pattern has been}%
\typeout{** loaded for the language `#1'. Using the pattern for}%
\typeout{** the default language instead.}%
\else
\language=\csname l@#1\endcsname
\fi
#2}}
\providecommand{\BIBdecl}{\relax}
\BIBdecl

\bibitem{shi2019face}
T.~Shi, Y.~Yuan, C.~Fan, Z.~Zou, Z.~Shi, and Y.~Liu, ``Face-to-parameter
  translation for game character auto-creation,'' in \emph{Proceedings of the
  IEEE/CVF International Conference on Computer Vision}, 2019, pp. 161--170.

\bibitem{shi2020neutral}
T.~Shi, Z.~Zou, X.~Song, Z.~Song, C.~Gu, C.~Fan, and Y.~Yuan, ``Neutral face
  game character auto-creation via pokerface-gan,'' in \emph{Proceedings of the
  28th ACM International Conference on Multimedia}, 2020, pp. 3201--3209.

\bibitem{song2020accurate}
S.~L. Song, W.~Shi, and M.~Reed, ``Accurate face rig approximation with deep
  differential subspace reconstruction,'' \emph{ACM Transactions on Graphics
  (TOG)}, vol.~39, no.~4, pp. 34--1, 2020.

\bibitem{lightstage}
\url{https://vgl.ict.usc.edu/LightStages/}.

\bibitem{zhou20183d}
J.~Zhou, H.-T. Wu, Z.~Liu, X.~Tong, and B.~Guo, ``3d cartoon face rigging from
  sparse examples,'' \emph{The Visual Computer}, vol.~34, no.~9, pp.
  1177--1187, 2018.

\bibitem{hong2022headnerf}
Y.~Hong, B.~Peng, H.~Xiao, L.~Liu, and J.~Zhang, ``Headnerf: A real-time
  nerf-based parametric head model,'' in \emph{Proceedings of the IEEE/CVF
  Conference on Computer Vision and Pattern Recognition}, 2022, pp.
  20\,374--20\,384.

\bibitem{anbarjafari20173d}
G.~Anbarjafari, R.~E. Haamer, I.~L{\"{u}}si, T.~Tikk, and L.~Valgma, ``3d face
  reconstruction with region based best fit blending using mobile phone for
  virtual reality based social media,'' \emph{Bulletin of the Polish Academy of
  Sciences. Technical Sciences}, vol.~67, no.~1, pp. 125--132, 2019.

\bibitem{lifkooee2018image}
M.~Z. Lifkooee, C.~Liu, M.~Li, and X.~Li, ``Image-based human character
  modeling and reconstruction for virtual reality exposure therapy,'' in
  \emph{International Conference on Computer Science \& Education (ICCSE)},
  2018, pp. 1--5.

\bibitem{DBLP:journals/tmm/LouWNHMWY20}
J.~Lou, Y.~Wang, C.~Nduka, M.~Hamedi, I.~Mavridou, F.-Y. Wang, and H.~Yu,
  ``Realistic facial expression reconstruction for vr hmd users,'' \emph{IEEE
  Transactions on Multimedia}, vol.~22, no.~3, pp. 730--743, 2019.

\bibitem{9537697}
X.~Wang, Y.~Guo, Z.~Yang, and J.~Zhang, ``Prior-guided multi-view 3d head
  reconstruction,'' \emph{IEEE Transactions on Multimedia}, pp. 1--1, 2021.

\bibitem{zhang2022video}
L.~Zhang, C.~Zeng, Q.~Zhang, H.~Lin, R.~Cao, W.~Yang, L.~Xu, and J.~Yu,
  ``Video-driven neural physically-based facial asset for production,''
  \emph{CoRR}, vol. abs/2202.05592, 2022.

\bibitem{li2021topologically}
T.~Li, S.~Liu, T.~Bolkart, J.~Liu, H.~Li, and Y.~Zhao, ``Topologically
  consistent multi-view face inference using volumetric sampling,'' in
  \emph{Proceedings of the IEEE/CVF International Conference on Computer
  Vision}, 2021, pp. 3824--3834.

\bibitem{wang2020lightweight}
X.~Wang, Y.~Guo, B.~Deng, and J.~Zhang, ``Lightweight photometric stereo for
  facial details recovery,'' in \emph{Proceedings of the IEEE/CVF Conference on
  Computer Vision and Pattern Recognition}, 2020, pp. 740--749.

\bibitem{cao2018sparse}
X.~Cao, Z.~Chen, A.~Chen, X.~Chen, S.~Li, and J.~Yu, ``Sparse photometric 3d
  face reconstruction guided by morphable models,'' in \emph{Proceedings of the
  IEEE Conference on Computer Vision and Pattern Recognition}, 2018, pp.
  4635--4644.

\bibitem{chen20203d}
Z.~Chen, Y.~Ji, M.~Zhou, S.~B. Kang, and J.~Yu, ``3d face reconstruction using
  color photometric stereo with uncalibrated near point lights,'' in \emph{2020
  IEEE International Conference on Computational Photography (ICCP)}, 2020, pp.
  1--12.

\bibitem{debevec2000acquiring}
P.~Debevec, T.~Hawkins, C.~Tchou, H.-P. Duiker, W.~Sarokin, and M.~Sagar,
  ``Acquiring the reflectance field of a human face,'' in \emph{Proceedings of
  the 27th annual conference on Computer graphics and interactive techniques},
  2000, pp. 145--156.

\bibitem{jensen2021benchmark}
S.~H.~N. Jensen, M.~E.~B. Doest, H.~Aan{\ae}s, and A.~Del~Bue, ``A benchmark
  and evaluation of non-rigid structure from motion,'' \emph{International
  Journal of Computer Vision}, vol. 129, no.~4, pp. 882--899, 2021.

\bibitem{kong2016prior}
C.~Kong and S.~Lucey, ``Prior-less compressible structure from motion,'' in
  \emph{Proceedings of the IEEE Conference on Computer Vision and Pattern
  Recognition}, 2016, pp. 4123--4131.

\bibitem{dai2014simple}
Y.~Dai, H.~Li, and M.~He, ``A simple prior-free method for non-rigid
  structure-from-motion factorization,'' \emph{International Journal of
  Computer Vision}, vol. 107, no.~2, pp. 101--122, 2014.

\bibitem{zhu2014complex}
Y.~Zhu, D.~Huang, F.~De~La~Torre, and S.~Lucey, ``Complex non-rigid motion 3d
  reconstruction by union of subspaces,'' in \emph{Proceedings of the IEEE
  conference on computer vision and pattern recognition}, 2014, pp. 1542--1549.

\bibitem{Sidhu2020}
V.~Sidhu, E.~Tretschk, V.~Golyanik, A.~Agudo, and C.~Theobalt, ``Neural dense
  non-rigid structure from motion with latent space constraints,'' in
  \emph{Computer Vision - {ECCV} 2020 - 16th European Conference, Glasgow, UK,
  August 23-28, 2020, Proceedings, Part {XVI}}, ser. Lecture Notes in Computer
  Science, vol. 12361, 2020, pp. 204--222.

\bibitem{ansari2017scalable}
M.~D. Ansari, V.~Golyanik, and D.~Stricker, ``Scalable dense monocular surface
  reconstruction,'' in \emph{2017 International Conference on 3D Vision (3DV)},
  2017, pp. 78--87.

\bibitem{golyanik2019consolidating}
V.~Golyanik, A.~Jonas, and D.~Stricker, ``Consolidating segmentwise non-rigid
  structure from motion,'' in \emph{2019 16th International Conference on
  Machine Vision Applications (MVA)}, 2019, pp. 1--6.

\bibitem{golyanik2017dense}
V.~Golyanik and D.~Stricker, ``Dense batch non-rigid structure from motion in a
  second,'' in \emph{2017 IEEE Winter Conference on Applications of Computer
  Vision (WACV)}, 2017, pp. 254--263.

\bibitem{bai2020deep}
Z.~Bai, Z.~Cui, J.~A. Rahim, X.~Liu, and P.~Tan, ``Deep facial non-rigid
  multi-view stereo,'' in \emph{Proceedings of the IEEE/CVF Conference on
  Computer Vision and Pattern Recognition}, 2020, pp. 5850--5860.

\bibitem{bai2021riggable}
Z.~Bai, Z.~Cui, X.~Liu, and P.~Tan, ``Riggable 3d face reconstruction via
  in-network optimization,'' in \emph{Proceedings of the IEEE/CVF Conference on
  Computer Vision and Pattern Recognition}, 2021, pp. 6216--6225.

\bibitem{grassal2022neural}
P.-W. Grassal, M.~Prinzler, T.~Leistner, C.~Rother, M.~Nie{\ss}ner, and
  J.~Thies, ``Neural head avatars from monocular rgb videos,'' in
  \emph{Proceedings of the IEEE/CVF Conference on Computer Vision and Pattern
  Recognition}, 2022, pp. 18\,653--18\,664.

\bibitem{mildenhall2020nerf}
B.~Mildenhall, P.~P. Srinivasan, M.~Tancik, J.~T. Barron, R.~Ramamoorthi, and
  R.~Ng, ``Nerf: Representing scenes as neural radiance fields for view
  synthesis,'' in \emph{European conference on computer vision}, 2020, pp.
  405--421.

\bibitem{tretschk2021non}
E.~Tretschk, A.~Tewari, V.~Golyanik, M.~Zollh{\"o}fer, C.~Lassner, and
  C.~Theobalt, ``Non-rigid neural radiance fields: Reconstruction and novel
  view synthesis of a dynamic scene from monocular video,'' in
  \emph{Proceedings of the IEEE/CVF International Conference on Computer
  Vision}, 2021, pp. 12\,959--12\,970.

\bibitem{pumarola2021d}
A.~Pumarola, E.~Corona, G.~Pons-Moll, and F.~Moreno-Noguer, ``D-nerf: Neural
  radiance fields for dynamic scenes,'' in \emph{Proceedings of the IEEE/CVF
  Conference on Computer Vision and Pattern Recognition}, 2021, pp.
  10\,318--10\,327.

\bibitem{park2021hypernerf}
K.~Park, U.~Sinha, P.~Hedman, J.~T. Barron, S.~Bouaziz, D.~B. Goldman,
  R.~Martin{-}Brualla, and S.~M. Seitz, ``Hypernerf: a higher-dimensional
  representation for topologically varying neural radiance fields,''
  \emph{{ACM} Trans. Graph.}, vol.~40, no.~6, pp. 238:1--238:12, 2021.

\bibitem{FLAME:SiggraphAsia2017}
T.~Li, T.~Bolkart, M.~J. Black, H.~Li, and J.~Romero, ``Learning a model of
  facial shape and expression from {4D} scans,'' \emph{ACM Transactions on
  Graphics, (Proc. SIGGRAPH Asia)}, vol.~36, no.~6, pp. 194:1--194:17, 2017.

\bibitem{lee2020maskgan}
C.-H. Lee, Z.~Liu, L.~Wu, and P.~Luo, ``Maskgan: Towards diverse and
  interactive facial image manipulation,'' in \emph{Proceedings of the IEEE/CVF
  Conference on Computer Vision and Pattern Recognition}, 2020, pp. 5549--5558.

\bibitem{wang2021neus}
P.~Wang, L.~Liu, Y.~Liu, C.~Theobalt, T.~Komura, and W.~Wang, ``Neus: Learning
  neural implicit surfaces by volume rendering for multi-view reconstruction,''
  in \emph{Advances in Neural Information Processing Systems 34: Annual
  Conference on Neural Information Processing Systems 2021, NeurIPS 2021},
  2021, pp. 27\,171--27\,183.

\bibitem{lorensen1987marching}
W.~E. Lorensen and H.~E. Cline, ``Marching cubes: A high resolution 3d surface
  construction algorithm,'' \emph{ACM siggraph computer graphics}, vol.~21,
  no.~4, pp. 163--169, 1987.

\bibitem{bregler2000recovering}
C.~Bregler, A.~Hertzmann, and H.~Biermann, ``Recovering non-rigid 3d shape from
  image streams,'' in \emph{Proceedings IEEE Conference on Computer Vision and
  Pattern Recognition. CVPR 2000 (Cat. No. PR00662)}, vol.~2, 2000, pp.
  690--696.

\bibitem{agudo2017global}
A.~Agudo and F.~Moreno-Noguer, ``Global model with local interpretation for
  dynamic shape reconstruction,'' in \emph{2017 IEEE Winter Conference on
  Applications of Computer Vision (WACV)}, 2017, pp. 264--272.

\bibitem{garg2013dense}
R.~Garg, A.~Roussos, and L.~Agapito, ``Dense variational reconstruction of
  non-rigid surfaces from monocular video,'' in \emph{Proceedings of the IEEE
  Conference on computer vision and pattern recognition}, 2013, pp. 1272--1279.

\bibitem{kumar2018scalable}
S.~Kumar, A.~Cherian, Y.~Dai, and H.~Li, ``Scalable dense non-rigid
  structure-from-motion: A grassmannian perspective,'' in \emph{Proceedings of
  the IEEE Conference on Computer Vision and Pattern Recognition}, 2018, pp.
  254--263.

\bibitem{kong2019deep}
C.~Kong and S.~Lucey, ``Deep non-rigid structure from motion,'' in
  \emph{Proceedings of the IEEE/CVF International Conference on Computer
  Vision}, 2019, pp. 1558--1567.

\bibitem{novotny2019c3dpo}
D.~Novotny, N.~Ravi, B.~Graham, N.~Neverova, and A.~Vedaldi, ``C3dpo: Canonical
  3d pose networks for non-rigid structure from motion,'' in \emph{Proceedings
  of the IEEE/CVF International Conference on Computer Vision}, 2019, pp.
  7688--7697.

\bibitem{takmaz2021unsupervised}
A.~Takmaz, D.~P. Paudel, T.~Probst, A.~Chhatkuli, M.~R. Oswald, and
  L.~Van~Gool, ``Unsupervised monocular depth reconstruction of non-rigid
  scenes,'' in \emph{2021 International Conference on 3D Vision (3DV)}, 2021,
  pp. 825--836.

\bibitem{lou2021real}
J.~Lou, X.~Cai, J.~Dong, and H.~Yu, ``Real-time 3d facial tracking via cascaded
  compositional learning,'' \emph{IEEE Transactions on Image Processing},
  vol.~30, pp. 3844--3857, 2021.

\bibitem{wu2021f3a}
X.~Wu, Q.~Zhang, Y.~Wu, H.~Wang, S.~Li, L.~Sun, and X.~Li, ``F$^3$a-gan: Facial
  flow for face animation with generative adversarial networks,'' \emph{IEEE
  Transactions on Image Processing}, vol.~30, pp. 8658--8670, 2021.

\bibitem{garrido2013reconstructing}
P.~Garrido, L.~Valgaerts, C.~Wu, and C.~Theobalt, ``Reconstructing detailed
  dynamic face geometry from monocular video.'' \emph{ACM Trans. Graph.},
  vol.~32, no.~6, pp. 158--1, 2013.

\bibitem{garrido2016reconstruction}
P.~Garrido, M.~Zollh{\"o}fer, D.~Casas, L.~Valgaerts, K.~Varanasi,
  P.~P{\'e}rez, and C.~Theobalt, ``Reconstruction of personalized 3d face rigs
  from monocular video,'' \emph{ACM Transactions on Graphics (TOG)}, vol.~35,
  no.~3, pp. 1--15, 2016.

\bibitem{blanz1999morphable}
V.~Blanz and T.~Vetter, ``A morphable model for the synthesis of 3d faces,'' in
  \emph{Proceedings of the 26th annual conference on Computer graphics and
  interactive techniques}, 1999, pp. 187--194.

\bibitem{paysan20093d}
P.~Paysan, R.~Knothe, B.~Amberg, S.~Romdhani, and T.~Vetter, ``A 3d face model
  for pose and illumination invariant face recognition,'' in \emph{2009 sixth
  IEEE international conference on advanced video and signal based
  surveillance}, 2009, pp. 296--301.

\bibitem{booth20163d}
J.~Booth, A.~Roussos, S.~Zafeiriou, A.~Ponniah, and D.~Dunaway, ``A 3d
  morphable model learnt from 10,000 faces,'' in \emph{Proceedings of the IEEE
  conference on computer vision and pattern recognition}, 2016, pp. 5543--5552.

\bibitem{deng2019accurate}
Y.~Deng, J.~Yang, S.~Xu, D.~Chen, Y.~Jia, and X.~Tong, ``Accurate 3d face
  reconstruction with weakly-supervised learning: From single image to image
  set,'' in \emph{{IEEE} Conference on Computer Vision and Pattern Recognition
  Workshops}, 2019, pp. 285--295.

\bibitem{guo2018cnn}
Y.~Guo, J.~Cai, B.~Jiang, J.~Zheng \emph{et~al.}, ``Cnn-based real-time dense
  face reconstruction with inverse-rendered photo-realistic face images,''
  \emph{IEEE transactions on pattern analysis and machine intelligence},
  vol.~41, no.~6, pp. 1294--1307, 2018.

\bibitem{guo20213d}
Y.~Guo, L.~Cai, and J.~Zhang, ``3d face from x: Learning face shape from
  diverse sources,'' \emph{IEEE Transactions on Image Processing}, vol.~30, pp.
  3815--3827, 2021.

\bibitem{zheng2021avatar}
Y.~Zheng, V.~F. Abrevaya, M.~C. B{\"u}hler, X.~Chen, M.~J. Black, and
  O.~Hilliges, ``Im avatar: Implicit morphable head avatars from videos,'' in
  \emph{Proceedings of the IEEE/CVF Conference on Computer Vision and Pattern
  Recognition}, 2022, pp. 13\,545--13\,555.

\bibitem{jiang2022selfrecon}
B.~Jiang, Y.~Hong, H.~Bao, and J.~Zhang, ``Selfrecon: Self reconstruction your
  digital avatar from monocular video,'' in \emph{Proceedings of the IEEE/CVF
  Conference on Computer Vision and Pattern Recognition}, 2022, pp. 5605--5615.

\bibitem{gafni2021dynamic}
G.~Gafni, J.~Thies, M.~Zollhofer, and M.~Nie{\ss}ner, ``Dynamic neural radiance
  fields for monocular 4d facial avatar reconstruction,'' in \emph{Proceedings
  of the IEEE/CVF Conference on Computer Vision and Pattern Recognition}, 2021,
  pp. 8649--8658.

\bibitem{flametexture}
\url{https://github.com/HavenFeng/photometric_optimization.git}.

\bibitem{karras2019style}
T.~Karras, S.~Laine, and T.~Aila, ``A style-based generator architecture for
  generative adversarial networks,'' in \emph{Proceedings of the IEEE/CVF
  conference on computer vision and pattern recognition}, 2019, pp. 4401--4410.

\bibitem{bulat2017far}
A.~Bulat and G.~Tzimiropoulos, ``How far are we from solving the 2d \& 3d face
  alignment problem?(and a dataset of 230,000 3d facial landmarks),'' in
  \emph{Proceedings of the IEEE International Conference on Computer Vision},
  2017, pp. 1021--1030.

\bibitem{wang2018high}
T.-C. Wang, M.-Y. Liu, J.-Y. Zhu, A.~Tao, J.~Kautz, and B.~Catanzaro,
  ``High-resolution image synthesis and semantic manipulation with conditional
  gans,'' in \emph{Proceedings of the IEEE conference on computer vision and
  pattern recognition}, 2018, pp. 8798--8807.

\bibitem{karras2017progressive}
T.~Karras, T.~Aila, S.~Laine, and J.~Lehtinen, ``Progressive growing of gans
  for improved quality, stability, and variation,'' in \emph{6th International
  Conference on Learning Representations, {ICLR} 2018}, 2018.

\bibitem{yariv2020multiview}
L.~Yariv, Y.~Kasten, D.~Moran, M.~Galun, M.~Atzmon, B.~Ronen, and Y.~Lipman,
  ``Multiview neural surface reconstruction by disentangling geometry and
  appearance,'' \emph{Advances in Neural Information Processing Systems},
  vol.~33, pp. 2492--2502, 2020.

\bibitem{gropp2020implicit}
A.~Gropp, L.~Yariv, N.~Haim, M.~Atzmon, and Y.~Lipman, ``Implicit geometric
  regularization for learning shapes,'' in \emph{Proceedings of the 37th
  International Conference on Machine Learning, {ICML} 2020}, ser. Proceedings
  of Machine Learning Research, vol. 119, 2020, pp. 3789--3799.

\bibitem{youtube}
\url{https://www.youtube.com/watch?v=a-_FuwTkFhI}.

\bibitem{tancik2020fourier}
M.~Tancik, P.~Srinivasan, B.~Mildenhall, S.~Fridovich-Keil, N.~Raghavan,
  U.~Singhal, R.~Ramamoorthi, J.~Barron, and R.~Ng, ``Fourier features let
  networks learn high frequency functions in low dimensional domains,''
  \emph{Advances in Neural Information Processing Systems}, vol.~33, pp.
  7537--7547, 2020.

\bibitem{park2021nerfies}
K.~Park, U.~Sinha, J.~T. Barron, S.~Bouaziz, D.~B. Goldman, S.~M. Seitz, and
  R.~Martin-Brualla, ``Nerfies: Deformable neural radiance fields,'' in
  \emph{Proceedings of the IEEE/CVF International Conference on Computer
  Vision}, 2021, pp. 5865--5874.

\bibitem{paszke2017automatic}
A.~Paszke, S.~Gross, S.~Chintala, G.~Chanan, E.~Yang, Z.~DeVito, Z.~Lin,
  A.~Desmaison, L.~Antiga, and A.~Lerer, ``Automatic differentiation in
  pytorch,'' 2017.

\bibitem{kingma2014adam}
D.~P. Kingma and J.~Ba, ``Adam: {A} method for stochastic optimization,'' in
  \emph{3rd International Conference on Learning Representations, {ICLR} 2015},
  2015.

\bibitem{feng2018prn}
Y.~Feng, F.~Wu, X.~Shao, Y.~Wang, and X.~Zhou, ``Joint 3d face reconstruction
  and dense alignment with position map regression network,'' in \emph{Computer
  Vision - {ECCV} 2018 - 15th European Conference, Munich, Germany, September
  8-14, 2018, Proceedings, Part {XIV}}, ser. Lecture Notes in Computer Science,
  vol. 11218.\hskip 1em plus 0.5em minus 0.4em\relax Springer, 2018, pp.
  557--574.

\bibitem{guo2020towards}
J.~Guo, X.~Zhu, Y.~Yang, F.~Yang, Z.~Lei, and S.~Z. Li, ``Towards fast,
  accurate and stable 3d dense face alignment,'' in \emph{Computer Vision -
  {ECCV} 2020 - 16th European Conference, Glasgow, UK, August 23-28, 2020,
  Proceedings, Part {XIX}}, ser. Lecture Notes in Computer Science, vol.
  12364.\hskip 1em plus 0.5em minus 0.4em\relax Springer, 2020, pp. 152--168.

\bibitem{3ddfa_cleardusk}
J.~Guo, X.~Zhu, and Z.~Lei, ``3ddfa,''
  \url{https://github.com/cleardusk/3DDFA}, 2018.

\bibitem{dib2021practical}
A.~Dib, G.~Bharaj, J.~Ahn, C.~Th{\'e}bault, P.~Gosselin, M.~Romeo, and
  L.~Chevallier, ``Practical face reconstruction via differentiable ray
  tracing,'' in \emph{Computer Graphics Forum}, vol.~40, no.~2, 2021, pp.
  153--164.

\bibitem{dib2021towards}
A.~Dib, C.~Thebault, J.~Ahn, P.-H. Gosselin, C.~Theobalt, and L.~Chevallier,
  ``Towards high fidelity monocular face reconstruction with rich reflectance
  using self-supervised learning and ray tracing,'' in \emph{Proceedings of the
  IEEE/CVF International Conference on Computer Vision}, 2021, pp.
  12\,819--12\,829.

\bibitem{dib2022s2f2}
A.~Dib, J.~Ahn, C.~Thebault, P.~Gosselin, and L.~Chevallier, ``{S2F2:}
  self-supervised high fidelity face reconstruction from monocular image,''
  \emph{CoRR}, vol. abs/2203.07732, 2022.

\bibitem{valgaerts2012lightweight}
L.~Valgaerts, C.~Wu, A.~Bruhn, H.-P. Seidel, and C.~Theobalt, ``Lightweight
  binocular facial performance capture under uncontrolled lighting.'' \emph{ACM
  Trans. Graph.}, vol.~31, no.~6, pp. 187--1, 2012.

\end{thebibliography}
}

% \begin{IEEEbiography} 
% [{\includegraphics[width=1in]{figures/authors/XueyingWang.png}}]{Xueying Wang} is a Ph.D. student at the School of Mathematical Sciences, University of Science and Technology of China. She obtained her bachelor's degree from Northeastern University in 2017. Her research interests include Computer Vision and Computer Graphics.
% \end{IEEEbiography}

% \begin{IEEEbiography} 
% [{\includegraphics[width=1in]{figures/authors/Juyong-Photo.jpg}}]{Juyong Zhang} is a professor in the School of Mathematical Sciences at the University of Science and Technology of China. He received his BS degree from University of Science and Technology of China in 2006, and Ph.D. degree from Nanyang Technological University, Singapore. His research interests include computer graphics, 3D computer vision and numerical optimization. He is currently an Associate Editor for IEEE Transaction on Multimedia and the Visual Computer Journal. 
% \end{IEEEbiography}

\end{document}